\documentclass[10pt,twocolumn,letterpaper]{article}
\usepackage{titlesec}
\usepackage{booktabs}
\usepackage{colortbl}
\usepackage{caption}
\usepackage{multirow}
\usepackage{algorithm}
\usepackage{algorithmic}
\usepackage{minitoc}
\usepackage{titletoc}
\usepackage[accsupp]{axessibility}  
\usepackage[pagenumbers]{iccv} 

\definecolor{iccvblue}{rgb}{0.21,0.49,0.74}
\definecolor{lightred}{RGB}{255, 230, 230} 
\definecolor{lightgreen}{RGB}{230, 255, 230} 
\definecolor{mediumred}{RGB}{255, 204, 204} 
\definecolor{mediumgreen}{RGB}{204, 255, 204} 
\definecolor{forestgreen}{RGB}{34,139,34}
\definecolor{deepyellow}{RGB}{184,134,11}
\definecolor{BrickRed}{RGB}{203,65,84} 
\usepackage[pagebackref,breaklinks,colorlinks,allcolors=iccvblue]{hyperref}

\titlespacing*{\subsection}
  {0pt}{2pt}{1pt}
\titlespacing*{\section}
  {0pt}{3pt}{2pt}
\titlespacing*{\subsubsection}
  {0pt}{1pt}{1pt}
\titlespacing*{\paragraph}{0pt}{0ex}{0.75em}
\def\paperID{8101} 
\def\confName{ICCV}
\def\confYear{2025}

\title{Evading Data Provenance in Deep Neural Networks}
\author{
Hongyu Zhu\textsuperscript{1}\thanks{Equal contribution. $^{\dag}$Corresponding author.}
\quad
Sichu Liang\textsuperscript{2}\footnotemark[1]
\quad
Wenwen Wang\textsuperscript{3}
\quad
Zhuomeng Zhang\textsuperscript{1}
\quad
Fangqi Li\textsuperscript{1}
\quad
Shi-Lin Wang\textsuperscript{1}\footnotemark[2]\\
\normalsize
\textsuperscript{1}Shanghai Jiao Tong University \quad
\textsuperscript{2}Southeast University \quad
\textsuperscript{3}Carnegie Mellon University\\
\normalsize
Code: \url{https://github.com/dbsxfz/EscapingDOV}
}

\begin{document}
\maketitle
\begin{abstract}
Modern over-parameterized deep models are highly data-dependent, with large scale general-purpose and domain-specific datasets serving as the bedrock for rapid advancements. However, many datasets are proprietary or contain sensitive information, making unrestricted model training problematic. In the open world where data thefts cannot be fully prevented, Dataset Ownership Verification (DOV) has emerged as a promising method to protect copyright by detecting unauthorized model training and tracing illicit activities. Due to its diversity and superior stealth, evading DOV is considered extremely challenging. However, this paper identifies that previous studies have relied on oversimplistic evasion attacks for evaluation, leading to a false sense of security. We introduce a unified evasion framework, in which a teacher model first learns from the copyright dataset and then transfers task-relevant yet identifier-independent domain knowledge to a surrogate student using an out-of-distribution (OOD) dataset as the intermediary. Leveraging Vision-Language Models and Large Language Models, we curate the most informative and reliable subsets from the OOD gallery set as the final transfer set, and propose selectively transferring task-oriented knowledge to achieve a better trade-off between generalization and evasion effectiveness. Experiments across diverse datasets covering eleven DOV methods demonstrate our approach simultaneously eliminates all copyright identifiers and significantly outperforms nine state-of-the-art evasion attacks in both generalization and effectiveness, with moderate computational overhead. As a proof of concept, we reveal key vulnerabilities in current DOV methods, highlighting the need for long-term development to enhance practicality. 
\end{abstract}    
\section{Introduction}
\label{sec:intro}
The success of modern deep learning hinges heavily on abundant datasets, spanning from large-scale vision and multi-modal resources such as ImageNet \cite{deng2009imagenet} and LAION-5B \cite{schuhmann2022laion} to datasets tailored for vertical industries like face recognition \cite{liu2015deep} and computational pathology \cite{yang2023medmnist}. Curated meticulously through extensive human effort for collection, cleaning, and labeling, these datasets are protected as intellectual property and are available solely through licensed access. Some are released for academic purposes with restrictions on commercial exploitation, while others are proprietary assets, strictly prohibiting third-party model training.
Moreover, personal data—such as photos shared on social platforms—is increasingly at risk of being scraped without consent \cite{nbcnews_facial_recognition}. Consequently, unauthorized model training on restricted data poses an escalating threat to both intellectual property and privacy rights.

Despite growing regulatory enforcement, like the General Data Protection Regulation (GDPR) \cite{GDPR2018}, preventing unauthorized training remains a formidable challenge. Every stage of the data supply chain is vulnerable to attacks that funnel data into illicit markets \cite{van2018plug}, including insider threats \cite{stolfo2012fog}, side-channel exploits \cite{hogan2023dbreach}, illegal mobile app access \cite{zuo2019does}, and leakage from pretrained models \cite{jin2021cafe,xu2020stealing}. High-profile cases of data misuse in model training, such as the Cambridge Analytica scandal \cite{WikipediaCambridgeAnalytica} and the Flickr Leakage \cite{nbcnews_facial_recognition}, have intensified public scrutiny.

Consequently, preventing unauthorized model training at its source—the dataset itself—is crucial. Unlearnable Examples \cite{huang2021unlearnable} introduce perturbations to prevent models from learning meaningful features. However, beyond their susceptibility to adaptive attacks \cite{sandoval2023can}, they entirely obstruct legitimate dataset use. As an alternative, post hoc verification for data provenance, known as \textbf{Dataset Ownership Verification} (DOV), has garnered substantial attention. Inspired by watermarks used to track duplication, remixing, or exploitation of multi-modal content \cite{cox1997secure}, the \textit{backdoor watermark} embeds carefully crafted "poison" samples selected by the dataset owner. Models trained on watermarked data exhibit predefined behaviors in response to backdoor triggers, enabling model-agnostic owner attribution \cite{li2023black, tang2023did}.

To mitigate the impact of backdoor on authorized training, recent techniques introduce \textit{non-poisoning watermarks} that adjust prediction confidence without inducing misclassification, employing hypothesis testing for ownership tracing \cite{sablayrolles2020radioactive, zou2022anti, wenger2024data}. Non-intrusive \textit{dataset fingerprint} have also emerged, allowing verification via elevated confidence on memorized training samples \cite{maini2021dataset, liu2022your}. DOV is widely regarded as an effective, if not the only, solution for tracking unauthorized training without disrupting legitimate usage \cite{li2022untargeted, guo2024domain}, with rapid advancements underway.

However, when deployed in the real world, the arms race between attackers and defenders forms an ongoing strategic game. Adversaries attempt to suppress verification behaviors to evade detection, while defenders continually strengthen verification mechanisms to reliably identify unauthorized training. Validating the robustness of DOV against evasion attempts is thus essential for practical deployment.
Despite substantial progress in DOV, advances in evasion attacks remain underdeveloped, relying on simple regularization techniques and generic adaptations from poisoning defenses like fine-tuning \cite{li2022untargeted, zou2022anti, liu2022your, guo2024domain}, which lack the versatility to counter diverse DOV techniques. Meanwhile, advanced DOV strategies, such as clean-label, invisible, or untargeted \textit{backdoor watermarks} provide exceptional flexibility; \textit{non-poisoning watermarks} and \textit{dataset fingerprints}, which avoid misclassification or require no dataset modifications, remain nearly undetectable. It is widely agreed that designing a universal attack capable of evading all forms of verification is nearly impossible \cite{maini2021dataset, wenger2024data}. This gap between the advancement of DOV and evasion techniques hinders reliable assessments.

In this paper we ask: is current DOV robust enough to preclude any evasion attempt? Our findings reveal that \textbf{prior evasion attacks are too weak, leading to a false sense of security}. Through analysis of verification behaviors, we identify shared characteristics: they are both \textit{exclusive} and \textit{subtle}, serving as unique markers that distinguish the protected dataset while remain incompatible with underlying semantic distribution of the main task.

\begin{figure*}[htbp]
\centerline{\includegraphics[width=1\linewidth]{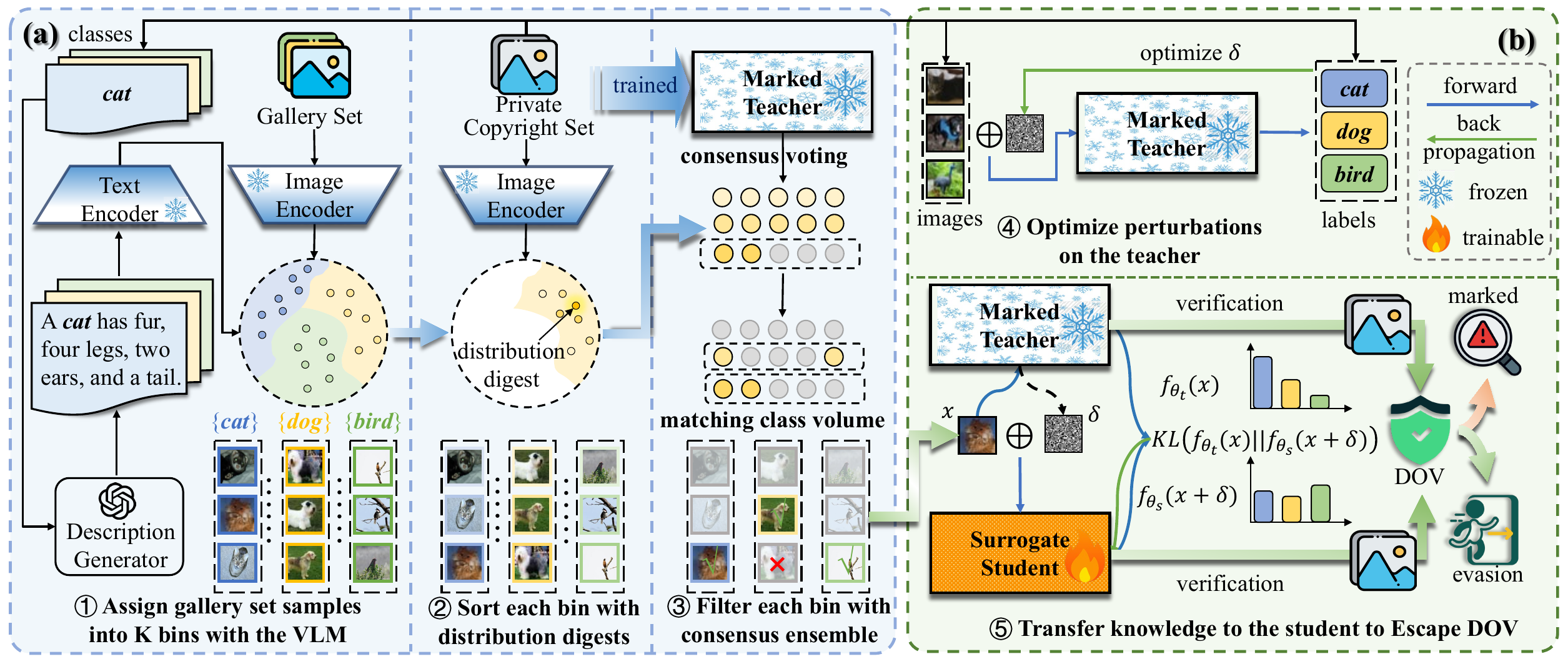}}  
\caption{\textbf{Overall Pipeline of Escaping DOV.} 
\textbf{(a) Transfer Set Curation} (\S \ref{sec:tsc}): 
Samples from the OOD gallery set are clustered via a VLM, ranked by distribution proximity, and filtered through consensus ensemble to curate an optimal transfer set.
\textbf{(b) Selective Knowledge Transfer} (\S \ref{sec:skt}): 
Promoting invariance to teacher's worst perturbations, facilitating task-oriented yet identifier-free knowledge transfer.
}
\label{fig:pipeline}
\end{figure*}

Building on this insight, we propose the first universal evasion framework, \textbf{Escaping DOV}, which requires no additional clean in-distribution data or assumptions about the verification process. In Escaping DOV, a teacher model is initially trained directly on the copyright dataset to fully absorb domain knowledge. Then, an OOD transfer set, unrelated to the protected data, serves as an intermediary to extract \textit{task-oriented} yet \textit{identifier-free} knowledge to a surrogate student for deployment. Intuitively, exclusive and subtle identifiers, acting as side-channel signals orthogonal to the task distribution, make it improbable for OOD samples to trigger verification behaviors. Thus, only essential, identifier-invariant task knowledge is likely to transfer to the student model.
To enhance Escaping DOV, we propose Transfer Set Curation, leveraging the unbiased knowledge embedded in large language models (LLMs) and vision-language models (VLMs) to retrieve the most \textit{informative} and \textit{reliable} samples from a general OOD gallery set. Additionally, we introduce Selective Knowledge Transfer to block suspicious verification behaviors, achieving better balance between generalization and evasion. The overall pipeline for Escaping DOV is shown in Figure~\ref{fig:pipeline}.

Our contributions are summarized as follows:
\begin{enumerate} 
\item We reveal that current DOV robustness evaluations are insufficiently rigorous. We introduce Escaping DOV, the first universal evasion framework that achieves simplicity, efficacy, and generalizability.
\item We propose Transfer Set Curation and Selective Knowledge Transfer to achieve balance between generalization and evasion with moderate overhead.
\item Experiments on diverse datasets and 11 DOV methods, alongside comparisons with 9 SOTA evasion methods, validate the efficacy of Escaping DOV, establishing it as a reliable benchmark for future DOV assessments.
\end{enumerate}
\section{Related Work}
\subsection{Dataset Ownership Verification}
\textit{Backdoor attacks} inject poisoned samples to associate predictions with a specific trigger \cite{gu2017identifying}, which induces misclassification whenever the trigger appears \cite{goldblum2022dataset}. When controlled by copyright owners, backdoors can also act as identifiers for model ownership \cite{adi2018turning}. Inspired by this, the straightforward \textit{backdoor watermark} embeds backdoor samples in the dataset. Models trained on this marked data produce a predefined label with high probability when evaluated on trigger samples \cite{li2023black}. The defender’s \textit{knowledge} of this secret prediction behavior serves as \textit{proof} of ownership.

Early backdoors link triggers to target classes by manipulating the labels of poisoned samples. However, due to the obvious nature of \textit{poisoned labels}, these samples are easily filtered through automatic data sanitization \cite{287222}. Clean-label backdoor watermark \cite{tang2023did} utilizes advanced poisoning techniques \cite{turner2019label} to enforce backdoor behavior without modifying labels. Later advancements further constrain modifications within an $L_\infty$ norm ball \cite{zeng2023narcissus}, making such \textit{clean-label, invisible} backdoor watermarks highly stealthy \cite{287222}. Recently, untargeted backdoor watermark \cite{li2022untargeted} induces non-deterministic misclassification without a fixed target class, enhancing resilience against post-training defenses such as trigger synthesis and output distribution analysis.

Advanced backdoor watermark provides a stealthy and resilient verification mechanism. However, it inevitably embeds harmful shortcuts \cite{zhu2024reliable}, problematic when authorized training is essential \cite{guo2024domain}.
\textit{Non-poisoning watermark} subtly embeds watermark features into the dataset without inducing misclassification. Verification is performed by measuring loss differences between watermarked and clean samples via hypothesis testing. Various mechanisms have been developed to generate stealthy watermarks: Radioactive data \cite{sablayrolles2020radioactive} and ML Auditor \cite{huang2024general} optimize $L_\infty$-perturbations, Anti-Neuron Watermarking \cite{zou2022anti} applies color transformations in hue space, Domain Watermark \cite{guo2024domain} converts samples into a hardly-generalizable domain, while Isotope \cite{wenger2024data} blends external features into the dataset.

Additionally, \textit{dataset fingerprint} utilizes intrinsic features for verification without modifying the dataset. Dataset inference \cite{maini2021dataset} analyzes the margins between training and external samples relative to the decision boundary, while MeFA \cite{liu2022your} aggregates membership inference across multiple samples. Although dataset fingerprints are non-intrusive and support post hoc protection for published datasets, they are susceptible to high false positives \cite{szyller2023conflicting, szyller2023robustness}.

\subsection{Evasion Attacks against Data Provenance}
Despite rapid progress in DOV, current evasion strategies rely on basic regularization and backdoor defenses like fine-tuning \cite{liu2018fine, li2022untargeted, guo2024domain}. \textbf{As they barely evade any DOV methods, we present their results in Appendix \textcolor{iccvblue}{C.1} and use SOTA backdoor defenses as baselines}, which are categorized by stage of application: (1) \textit{Data sanitization} removes suspicious samples \textit{pre-training} \cite{287222}; (2) \textit{Robust learning} mitigates poisoning effects \textit{during training} \cite{bu2023automatic, li2021neural}; (3) \textit{Backdoor unlearning} synthesizes triggers \textit{post-training} to neutralize the model \cite{zeng2022adversarial}; and (4) \textit{Input preprocessing} detects or purifies triggers \textit{during testing} \cite{shi2023black}. Together, these defenses span the model lifecycle; yet advanced invisible, clean-label, or untargeted backdoor watermarks can bypass most of them \cite{li2022untargeted, guo2024domain}.
Further, non-poisoning watermarks and non-intrusive fingerprints render backdoor defenses ineffective, making universal evasion extremely challenging. This work is the first to establish a unified evasion perspective, opening new directions for robust evaluation of DOV.
We discuss the contemporaneous works \cite{Zhu_2025_ECAI} and \cite{shao2025databench} that emerged at the time of acceptance in Appendix E.
\section{Escaping DOV}
\subsection{Overview}
In $K$-class image classification, an adversary gains unauthorized access to the copyright dataset $\mathcal{D}$ to train a model $f_{\theta}$, which maps images \( x \in \{0, 1, \dots, 255\}^{C \times W \times H} \) to predictions $y \in \mathbb{R}^K$. For data provenance, the defender employs a DOV method $\text{Verif}(f_{\theta}, \mathcal{D}, \mathcal{M})$, inspecting the output of the suspicious model $f_{\theta}$ on specific queries. This process yields a verification metric indicating the likelihood that $f_{\theta}$ was trained on $\mathcal{D}$, such as the Attack Success Rate for backdoor watermarks or the confidence for rejecting the null hypothesis in non-poisoning DOV. Here, $\mathcal{M}$ represents external materials (e.g., watermark trigger patterns) controlled by the defender to uniquely identify the dataset.

 DOV methods are flexible and diverse. Backdoor watermark achieves superior stealth through invisible, clean-label, or untargeted poison techniques. Non-poisoning watermarks avoid misclassification, while dataset fingerprints require no dataset modifications. Hence, direct sanitization or suppression of all potential verification behaviors becomes a formidable challenge; most evasion attempts that simply adapt poison defenses have consistently failed \cite{li2022untargeted, zou2022anti, guo2024domain, wenger2024data, maini2021dataset}. However, we identify a common trait among all DOV strategies: verification behaviors are both \textit{exclusive} and \textit{subtle}. On one hand, these behaviors must \textit{exclusively} belong to the copyright dataset to prevent false attribution to models trained on external datasets. On the other hand, they must be \textit{subtle} enough to remain concealed during data sanitization, ensuring they are rarely triggered by regular samples to preserve the generalization of models trained for authorized use.

Accordingly, we propose a universal evasion strategy: first, a teacher model $f_{\theta_t}$ is trained on the copyright dataset $\mathcal{D}$ to absorb task-domain knowledge, inevitably marked by DOV. Next, an unrelated OOD transfer set $\mathcal{T}$ serves as the medium to extract task-oriented yet identifier-invariant knowledge into a surrogate student $f_{\theta_s}$. Intuitively, the \textit{exclusive} and \textit{subtle} watermark triggers are absent in the natural samples of $\mathcal{T}$, preventing the activation and transmission of verification behaviors to the surrogate student. For fingerprints based on exaggerated confidence, the student only indirectly inherits knowledge from the marked teacher via the unrelated OOD data, significantly reducing excessive memorization associated with $\mathcal{D}$.

The overall objective of Escaping DOV can be formalized as simultaneously minimizing the generalization error on the task distribution and the verification metric:
\begin{equation}
\min_{\theta_s} \mathbb{E}_{(x, y) \in P(x, y)} [\mathcal{L}(f_{\theta_s}(x), y)] + \alpha \cdot \text{Verif}(f_{\theta_s}, \mathcal{D}, \mathcal{M})
\label{eq:escaping_dov_objective}
\end{equation}
Here, \(f_{\theta_s}\) represents the surrogate student ultimately deployed, \(P(x, y)\) the underlying distribution of samples \(x, y \in \mathcal{D}\), and \(\mathcal{L}\) the loss function measuring prediction discrepancies. The tuning factor \(\alpha\) balances the pursuit of generalization and evasion for \(f_{\theta_s}\). However, \(P(x, y)\) and the process of \(\text{Verif}(f_{\theta_s}, \mathcal{D}, \mathcal{M})\) are agnostic to the adversary and intractable during optimization. Therefore, we propose two modules to better achieve and balance these objectives: (1) For \textit{generalization}, we curate the most informative and reliable transfer set from a large-scale OOD gallery set to support high-performance and identifier-invariant knowledge extraction; (2) For \textit{evasion}, to prevent the transmission of suspicious verification knowledge via dark knowledge in soft labels, we propose selective knowledge transfer to filter out such knowledge and better trade off between the two objectives. Next, we elaborate on these two components.

\subsection{Transfer Set Curation}
\label{sec:tsc}
Our objective is to curate an optimal subset $\mathcal{T}$ from an OOD gallery set \( \mathcal{G} \) (e.g., ImageNet \cite{deng2009imagenet} or DataComp \cite{gadre2024datacomp}), selecting samples that are both \textit{informative} and \textit{reliable} to facilitate effective knowledge transfer. While this task may resemble classical core-set selection \cite{sener2018active, chen2021learning}, it presents significantly greater challenge: (1) Our algorithm operates in a fully OOD context, where \( \mathcal{G} \) and \( \mathcal{D} \) do not necessarily overlap in classes, and their distributions could vary substantially; (2) \( \mathcal{D} \) could include identifiers manipulated by the defender (e.g., trigger samples), and the teacher \( f_{\theta_t} \) trained directly on \( \mathcal{D} \) is already marked. Consequently, selection based on the unreliable \( \mathcal{D} \) and \( f_{\theta_t} \) inadvertently introduce samples from \( \mathcal{G} \) that activate verification behaviors, risking the induction of these behaviors in the surrogate student \( f_{\theta_s} \).

In this paper, we leverage a vision-language model (VLM) with unbiased pretrained knowledge and robust zero-shot capabilities (e.g., CLIP \cite{radford2021learning}) to facilitate reliable selection alongside the marked teacher \( f_{\theta_t} \). CLIP aligns visual and textual features in a shared embedding space, where the image encoder \( \phi_I(\cdot) \) maps an image \( x \) into \( \phi_I(x) \), and text templates with category names (e.g., "a photo of \{class name\}") \( c_1, \ldots, c_K \in \mathcal{D} \) are mapped by the text encoder \( \phi_t(\cdot) \) to \( \phi_t(c_1), \ldots, \phi_t(c_K) \). By calculating the cosine similarity between \( \phi_I(x) \) and \( \phi_t(c_i) \), each sample \( x \in \mathcal{G} \) can be uniquely assigned to a category \( t \) in \( \mathcal{D} \).

However, class-name-only text templates lack foreground semantics \cite{pratt2023does}, limiting the ability to distinguish target categories. While few-shot tuning allows the VLM to rapidly adapt to tasks within \( \mathcal{D} \) \cite{jia2022visual}, it risks being compromised by verification behaviors embedded in \( \mathcal{D} \) \cite{bai2024badclip}. Inspired by advances in zero-shot prompt learning \cite{pratt2023does}, we adjust the VLM with unbiased category semantics from a large language model (LLM). The LLM generates a description set \( \text{Desc}_{c_i} \) for each category \( c_i \) in \( \mathcal{D} \), averaging descriptions in the feature space to create representation prototypes for each category, enhancing the zero-shot capabilities of the VLM. 
Consequently, each \( x \in \mathcal{G} \) is assigned to class \( t \) in a zero-shot fashion as follows in Equation \ref{eq:assignment_process}:
\begin{equation}
t = \arg\max_{c_i} \frac{1}{|\text{Desc}_{c_i}|} \sum_{d \in \text{Desc}_{c_i}} \left(\text{sim}(\phi_I(x), \phi_I(d))\right) 
\label{eq:assignment_process}
\end{equation}
where \( \text{sim} \) denotes cosine similarity shown in Equation \ref{eq:cosine_similarity}:
\begin{equation}
\text{sim}(\phi_I(x), \phi_I(d)) = \frac{\phi_I(x) \cdot \phi_I(d)}{|\phi_I(x)| \cdot |\phi_I(d)|}
\label{eq:cosine_similarity}
\end{equation}
Although the VLM assigns all gallery set samples to categories in \( \mathcal{D} \), it lacks a criterion for the most \textit{informative} samples. To leverage the distribution from \( \mathcal{D} \) while avoiding biases from potential verification behaviors, we project all samples in \( \mathcal{D} \) into the feature space with \( \phi_I(\cdot) \) and calculate the density centroids \( \text{Cent}_t \) for each class \( t \in \{1, \ldots, K\} \). These centroids, acting as \textbf{distribution digests}, encapsulate the conditional distribution of \( \mathcal{D} \). Samples closer to \( \text{Cent}_t \) in feature space are prioritized for selection, a robust and straightforward criterion that is theoretically resistant to manipulation by a small proportion of outlier samples \cite{jia2022certified}.

We then establish a consensus-based ensemble for final transfer set curation: (1) The gallery set \( \mathcal{G} \) is divided by the LLM-enhanced VLM into \( K \) bins corresponding to classes in \( \mathcal{D} \). Within each bin \( \mathcal{G}_t \), (2) samples are ordered by proximity to distribution digest \( \text{Cent}_t \); (3) Only samples classified by the teacher \( f_{\theta_t} \) as \( t \) that are closest to \( \text{Cent}_t \) are selected, until the selected samples match the class count \( |\mathcal{D}_t| \). Illustrated in Figure \ref{fig:pipeline}, this process ensures all predictions on the transfer set \( \mathcal{T} \) by \( f_{\theta_t} \) are consistent with VLM, minimizing the risk of triggering verification behaviors while closely approximating the distribution in \( \mathcal{D} \).

\noindent \textbf{The Feature Bank.} The curation process involves projecting all gallery set samples into the feature space, which is the most time-consuming step. However, when the gallery set is potentially reused, this projection only needs to be performed once. The resulting features and indexes are saved in a \textbf{feature bank} for efficient reuse.

\subsection{Selective Knowledge Transfer}
\label{sec:skt}
The transfer set $\mathcal{T}$ serves as an intermediary for extraction of task-oriented and identifier-invariant knowledge. This process employs the standard knowledge distillation framework \cite{hinton2015distilling} with the following optimization objective:
\begin{equation}
\theta_s^* = \arg\min_{\theta_s} \mathcal{L}\left(\frac{f_{\theta_s}(x)}{\tau}, \frac{f_{\theta_t}(x)}{\tau}\right), \quad x \in \mathcal{T}
\label{eq:distillation}
\end{equation}
Here, \( \mathcal{L} \) measures prediction discrepancies, such as the Kullback-Leibler divergence, modulated by the temperature \( \tau \). This process effectively transfers knowledge despite potential misalignments between the transfer set $\mathcal{T}$ and the training distribution \cite{orekondy2019knockoff}. However, the soft labels \( f_{\theta_t}(x) \) contain dark knowledge \cite{furlanello2018born}, which inadvertently transmit verification behaviors. Thus, we introduce Selective Knowledge Transfer (SKT) to filter out suspicious verification knowledge during distillation.

The \textit{exclusive} and \textit{subtle} verification behaviors reflect unique biases inherent to the marked dataset \( \mathcal{D} \), distinguishing it within the task distribution. Preventing the student \( f_{\theta_s} \) from overfitting to biases limits DOV's ability to infer connections between \( f_{\theta_s} \) and \( \mathcal{D} \) through trigger-specific outputs or excessive confidence. Drawing on insights from shortcut learning \cite{geirhos2020shortcut}, we mitigate predictive shortcuts during the student's learning process, reducing reliance on spurious features. Since dataset biases typically induce universal shortcut behaviors, a straightforward solution is to apply universal adversarial training (UAT) on the student:
\begin{equation}
\theta_s^* = \arg\min_{\theta_s} \max_{\delta} \frac{1}{n} \sum_{x, y \in \textcolor{BrickRed}{\mathcal{T}}} \mathcal{L}(f_{\textcolor{BrickRed}{\theta_s}}(x+\delta), y) \quad \label{eq:UAT_s}
\end{equation}
UAT effectively suppresses atypical predictive behaviors in \(f_{\theta_s}\) induced by perturbations \(\delta\). However, the min-max problem involves a notoriously hard bi-level optimization, requiring iterative solutions within each iteration. Furthermore, the absence of ground-truth labels in the OOD transfer set \( \mathcal{T} \) necessitates predictions from the marked teacher as labels, introducing significant calibration errors.

Given the objective to mitigate biases specific to the copyrighted dataset \( \mathcal{D} \), and with all verification behaviors encapsulated in the teacher model, we propose using only the teacher model \( f_{\theta_t} \) to generate perturbations on \( \mathcal{D} \) as a practical approximation for the inner max-problem:
\begin{equation}
\delta^* = \arg\max_{\delta} \frac{1}{n} \sum_{x, y \in \textcolor{BrickRed}{\mathcal{D}}} \mathcal{L}(f_{\textcolor{BrickRed}{\theta_t}}(x+\delta), y) \quad \label{eq:UAT_t}
\end{equation}
This strategy decouples the bi-level optimization, rendering it asynchronously solvable. By pre-generating a \textbf{perturbation pool} \(\{ \delta_i \}_{i=1}^n\) in an offline manner with the teacher \(f_{\theta_t}\) and reusing it in the outer min-problem during distillation, computational overhead is significantly reduced.

To solve Equation \ref{eq:UAT_t}, the perturbation \(\delta\) is typically constrained by an \(L_p\) norm to preserve core image semantics \cite{madry2018towards}. However, without prior knowledge of the specific form of \(\delta\) associated with verification behaviors (e.g., watermark triggers), choosing an inappropriate norm undermine the efficacy of UAT in suppressing spurious features \cite{tramer2019adversarial}. Thus, we seek to diversify the teacher-generated perturbations beyond a single norm constraint.

We solve the max-problem using mini-batch stochastic gradient descent \cite{shafahi2020universal}, applying \(L_2\) constraints at each batch update to preserve gradient directions. After each iteration, we project the perturbation onto the norm constraint (\(L_0\), \(L_2\), or \(L_{\infty}\)) that maximizes the loss. This steepest descent projection \cite{maini2020adversarial} adaptively seeks the optimal norm constraint, facilitating the generation of perturbations closely resembling those employed by DOV. Moreover, extreme projections onto \(L_0\) or \(L_{\infty}\) norms approximate a convex hull spanning multiple norm constraints, broadening the range of potential perturbations \cite{croce2022adversarial}.

\begin{table*}[htb]
  \centering
  \caption{Escaping DOV on CIFAR-10 and Tiny ImageNet. {\color{mediumred}Red} indicates detection by DOV, while  {\color{mediumgreen}green} indicates successful evasion.}
  \resizebox{\linewidth}{!}{
    \begin{tabular}{c|cccc|cccc}
      \toprule[1.5pt]
      \multirow{2}[0]{*}{\textbf{DOV}} & \multicolumn{4}{c|}{\textbf{CIFAR-10}} & \multicolumn{4}{c}{\textbf{Tiny-Imagenet}} \\
     \cmidrule(r){2-5} \cmidrule(l){6-9}
      & \multicolumn{2}{c}{Vanilla} & \multicolumn{2}{c|}{Evasion} & \multicolumn{2}{c}{Vanilla} & \multicolumn{2}{c}{Evasion} \\
      \midrule
      \rowcolor[rgb]{.906, .902, .902} \textbf{Poisoning DOV}& ACC & VSR & ACC(↑) & VSR(↓) & ACC & VSR & ACC(↑) & VSR(↓) \\
      \midrule
      \textbf{Badnets} & 94.44 ± 0.74 & \cellcolor{lightred} 100.00 ± 0.00 & 93.46 ± 0.20 &\cellcolor{lightgreen} 1.36 ± 0.07 & 69.05 ± 0.25 & \cellcolor{lightred} 98.98 ± 0.11 & 66.48 ± 0.14 &\cellcolor{lightgreen} 1.48 ± 0.11 \\
       \textbf{UBW} & 93.98 ± 0.25 & \cellcolor{lightred} 95.54 ± 0.84 & 93.41 ± 0.19 &\cellcolor{lightgreen} 1.74 ± 0.06 & 66.37 ± 0.12 &\cellcolor{lightred} 75.56 ± 0.50 & 65.34 ± 0.08 &\cellcolor{lightgreen} 4.60 ± 0.11 \\
      \textbf{Label-Consistent} & 94.74 ± 0.54 &\cellcolor{lightred} 96.18 ± 2.70 & 93.19 ± 0.14 &\cellcolor{lightgreen} 3.74 ± 0.40 & 69.07 ± 0.24 &\cellcolor{lightred} 34.99 ± 1.53 & 66.46 ± 0.10 &\cellcolor{lightgreen} 3.51 ± 0.22 \\
       \textbf{Narcissus} & 94.76 ± 0.54 &\cellcolor{lightred} 87.34 ± 1.82 & 94.37 ± 0.18 &\cellcolor{lightgreen} 4.59 ± 0.64 & 68.99 ± 0.14 &\cellcolor{lightred} 57.49 ± 0.78 & 66.39 ± 0.18 &\cellcolor{lightgreen} 12.33 ± 6.45 \\
      \midrule
    \rowcolor[rgb]{.906, .902, .902} \textbf{Non-Poisoning DOV} & ACC  & p-value & ACC(↑) & p-value(↑) & ACC   & p-value & ACC(↑) & p-value(↑) \\
    \midrule
    \textbf{Radioactive Data} & 94.96 ± 0.47 & \cellcolor{lightred} 3.03e-03 & 94.07 ± 0.18 & \cellcolor{lightgreen} 9.45e-01 & 68.84 ± 0.30 & \cellcolor{lightred} 5.33e-03 & 66.64 ± 0.24 &\cellcolor{lightgreen} 4.94e-01 \\
     \textbf{ANW} & 94.66 ± 0.31 & \cellcolor{lightred} 1.37e-09 & 93.91 ± 0.28 &\cellcolor{lightgreen} 1.00e+00 & 68.75 ± 0.30 & \cellcolor{lightred} 5.57e-28 & 66.44 ± 0.13 &\cellcolor{lightgreen} 1.00e+00 \\
    \textbf{Domain Watermark} & 94.38 ± 0.53 &\cellcolor{lightred} 1.67e-22 & 93.90 ± 0.13 &\cellcolor{lightgreen} 1.00e+00 & 68.55 ± 0.39 &\cellcolor{lightred} 3.91e-16 & 66.38 ± 0.16 &\cellcolor{lightgreen} 1.00e+00 \\
     \textbf{Isotope} & 94.75 ± 0.28 &\cellcolor{lightred} 2.87e-03 & 93.99 ± 0.49 &\cellcolor{lightgreen} 2.84e-01 & 68.97 ± 0.37 &\cellcolor{lightred} 1.63e-03 & 66.76 ± 0.22 &\cellcolor{lightgreen} 1.66e-01 \\
     \textbf{ML Auditor}& 94.22 ± 0.17& \cellcolor{lightred}7.47e-05&93.45 ±  0.21&\cellcolor{lightgreen}7.83e-01&68.41 ± 0.28&\cellcolor{lightred}1.55e-06&65.35 ± 0.16&\cellcolor{lightgreen}1.00e+00\\
    \textbf{Dataset Inference} & 94.83 ± 0.47 &\cellcolor{lightred} 1.87e-03 & 93.97 ± 0.49 &\cellcolor{lightgreen} 4.76e-01 & 69.15 ± 0.34 &\cellcolor{lightred} 9.27e-14 & 66.06 ± 0.18 &\cellcolor{lightgreen} 4.47e-01 \\
     \textbf{MeFA} & 94.83 ± 0.47 &\cellcolor{lightred} 2.62e-14 & 93.93 ± 0.55 &\cellcolor{lightgreen} 1.00e+00 & 69.15 ± 0.34 &\cellcolor{lightred} 6.17e-27 & 66.22 ± 0.20 &\cellcolor{lightgreen} 9.99e-01 \\
    \bottomrule[1.5pt]
    \end{tabular}
  }
  \label{tab:escaping_advanced}
\end{table*}

Ideally, perturbation should preserve core image semantics without being limited to specific norms \cite{song2018constructing}. Generative models can achieve this by producing unrestricted adversarial perturbations \cite{zhu2024reliable}, though training such models on task-specific distributions is resource-intensive. Therefore, we propose a lightweight approach using image corruptions instead of generative models to produce unrestricted perturbations. Starting with the 15 corruptions defined in ImageNet-C \cite{hendrycks2018benchmarking}, we employ a genetic algorithm \cite{deb2002fast, mao2021composite} to select the combination and sequence of 
corruptions that solves Equation \ref{eq:UAT_t}, forming a corruption chain:
\begin{equation}
\max \mathcal{L}(f_{\theta_t}(\text{Corrupt}_{1:N}(x)), y), \quad x, y \in \mathcal{D}
\end{equation}
The perturbations and corruption chain generated with the teacher \(f_{\theta_t}\) on \(\mathcal{D}\) are collectively referred to as \(A(\cdot)\), leading to the final objective for Selective Knowledge Transfer:
\begin{equation}
\begin{split}
\arg\min_{\theta_s} &\; \mathcal{L}\left(\frac{f_{\theta_s}(x)}{\tau}, \frac{f_{\theta_t}(x)}{\tau}\right) \\
&+ \beta \cdot \mathcal{L}\left(\frac{f_{\theta_s}(A(x))}{\tau}, \frac{f_{\theta_t}(x)}{\tau}\right), \quad x \in \mathcal{T}
\end{split}
\end{equation}

The second term promotes the student's invariance to the worst-case perturbations from the teacher while encouraging predictive divergence to reduce bias toward the teacher. The tuning factor \(\beta\) balances generalization with evasion efficacy, applied as a sampling probability to selectively introduce perturbations or corruption chain to a subset of samples, adding minimal overhead.

\section{Experiments}
\subsection{Experimental Setup}
Following established DOV methods \cite{li2023black, zou2022anti, guo2024domain, maini2021dataset}, we adopt ResNet-18 \cite{he2016deep} for both teacher and students. For transfer set curation, we utilize MobileCLIP \cite{vasu2024mobileclip} and GPT-4o mini \cite{openai_gpt4o_mini} as lightweight VLM and LLM, respectively.

We assess backdoor watermarks with Verification Success Rate (VSR)-the probability of a trigger-induced misclassification to the target class. 
\textbf{A VSR above 30\% confirms successful verification} \cite{zhu2024reliable}, indicating unauthorized training.
Non-poisoning watermarks and fingerprints, collectively termed non-poisoning DOV, are evaluated via the p-value from a one-tailed T-test. The p-value represents the confidence level that the model was \textbf{not} trained on the copyright dataset; \textbf{lower p-values indicate successful verification}. The common threshold for p-values is set at 0.01 \cite{maini2021dataset}, equating to a 99\% confidence level of unauthorized training. 
Experiments were repeated 3 times, reporting mean \( \pm \) std for ACC and VSR, and harmonic mean for p-values.

\subsection{Escaping Advanced DOV Methods}
\begin{table*}[htb]
  \centering
  \caption{Comparison of Escaping DOV with Evasion Attacks. {\color{mediumred}Red} indicates detection by DOV, {\color{mediumgreen}green} indicates successful evasion.}
  \resizebox{\linewidth}{!}{
    \begin{tabular}{c|cccccccc}
    \toprule[1.5pt]
       \multirow{2}[0]{*}{\textbf{Method}}  & \multicolumn{2}{c}{\textbf{Badnets}} & \multicolumn{2}{c}{\textbf{Narcissus}} & \multicolumn{2}{c}{\textbf{Isotope}} & \multicolumn{2}{c}{\textbf{Dataset Inference}}   \\
       \cmidrule(r){2-3} \cmidrule(lr){4-5} \cmidrule(lr){6-7} \cmidrule(l){8-9}
          & \cellcolor[rgb]{.906, .902, .902}ACC(↑) &\cellcolor[rgb]{.906, .902, .902} VSR(↓) &\cellcolor[rgb]{.906, .902, .902} ACC(↑) &\cellcolor[rgb]{.906, .902, .902} VSR(↓) &\cellcolor[rgb]{.906, .902, .902} ACC(↑) &\cellcolor[rgb]{.906, .902, .902} p-value(↑) &\cellcolor[rgb]{.906, .902, .902} ACC(↑) &\cellcolor[rgb]{.906, .902, .902} p-value(↑) \\
    \midrule
    \textbf{Fine-pruning} & 86.21 ± 0.16 & \cellcolor{lightred} 99.37 ± 0.27 & 86.58 ± 0.15 & \cellcolor{lightred} 43.75 ± 14.41 & 86.78 ± 0.12 & \cellcolor{lightgreen}0.0532  & 86.88 ± 0.48 &\cellcolor{lightred} 0.0095  \\
     \textbf{Meta-Sift} & 88.68 ± 0.45 & \cellcolor{lightred} 99.70 ± 0.32 & 88.64 ± 0.62 &\cellcolor{lightred} 49.40 ± 8.26 & 88.56 ± 0.47 &\cellcolor{lightgreen} 0.0264  & 88.07 ± 0.21 &\cellcolor{lightgreen} 0.0963  \\
    \textbf{Differential Privacy} & 90.49 ± 1.20 &\cellcolor{lightred} 76.70 ± 4.62 & 87.70 ± 0.68 &\cellcolor{lightred} 72.77 ± 3.91 & 88.69 ± 2.61 & \cellcolor{lightgreen}0.0678  & 89.64 ± 4.86 &\cellcolor{lightgreen} 0.0440  \\
     \textbf{I-BAU} & 89.09 ± 0.25 &\cellcolor{lightgreen} 14.80 ± 0.46 & 88.76 ± 1.94 &\cellcolor{lightred} 72.74 ± 2.17 & 87.07 ± 2.59 &\cellcolor{lightgreen} 0.1552  & 89.93 ± 0.68 &\cellcolor{lightgreen} 0.0355  \\
    \textbf{ZIP} & 82.44 ± 0.10 &\cellcolor{lightred} 79.88 ± 0.39 & 83.51 ± 0.37 &\cellcolor{lightred} 53.21 ± 4.47 & 82.97 ± 0.42 &\cellcolor{lightred} 0.0098  & 83.21 ± 0.90 &\cellcolor{lightgreen} 0.1050  \\
    \textbf{NAD} & 89.86 ± 0.37 &\cellcolor{lightgreen} 3.11 ± 0.51  & 91.04 ± 0.36 &\cellcolor{lightred} 75.41 ± 5.12 & 91.77 ± 0.45 & \cellcolor{lightgreen} 0.0961 &  92.12 ± 0.34 &\cellcolor{lightred} 0.0014 \\
     \textbf{BCU} & 92.75 ± 0.05 &\cellcolor{lightgreen} 1.56 ± 0.25 & 92.33 ± 0.11 &\cellcolor{lightred} 61.72 ± 6.43 & 92.77 ± 0.36 &\cellcolor{lightgreen} 0.1622  & 93.09 ± 0.18 &\cellcolor{lightgreen} 0.0773  \\
    \textbf{ABD} & 84.94 ± 3.04 &\cellcolor{lightgreen} 7.07 ± 3.15 & 85.42 ± 2.10 &\cellcolor{lightgreen} 22.10 ± 1.15 & 86.29 ± 3.46 &\cellcolor{lightred} 0.0066  & 84.87 ± 3.02 &\cellcolor{lightgreen} 0.3696  \\
     \textbf{IPRemoval} & 83.96 ± 0.23 &\cellcolor{lightgreen} 8.09 ± 0.50 & 85.06 ± 0.11 &\cellcolor{lightred} 55.97 ± 2.14 & 84.64 ± 0.06 &\cellcolor{lightgreen} 0.1916  & 85.28 ± 0.05 & \cellcolor{lightgreen} 0.1322  \\
    \textbf{Escaping DOV (Ours)} & \textbf{93.46 ± 0.20} &\cellcolor{lightgreen} \textbf{1.36 ± 0.07} & \textbf{94.37 ± 0.18} & \cellcolor{lightgreen} \textbf{4.59 ± 0.64} & \textbf{93.99 ± 0.49} &\cellcolor{lightgreen} \textbf{0.2845} & \textbf{93.97 ± 0.49} & \cellcolor{lightgreen} \textbf{0.4759} \\
    \bottomrule[1.5pt]
    \end{tabular}%
    }
  \label{tab:comparison}
\end{table*}%

We evaluate Escaping DOV on natural image datasets commonly used as DOV benchmarks, including CIFAR-10 \cite{krizhevsky2009learning} and the 200-class Tiny ImageNet \cite{le2015tiny}. \textbf{We further consider six datasets with significant distribution shifts from the gallery set} (e.g., facial recognition and medical diagnosis), with a summary of results in Section \ref{sec:domain-specific} and detailed findings in Appendix \textcolor{iccvblue}{C.3}. 
For the gallery set, we use LSVRC-2012 \cite{deng2009imagenet}. Given overlapping samples between Tiny ImageNet and LSVRC-2012, we creat a challenging setting by \textit{removing all overlapping classes from LSVRC-2012 to produce a purely OOD gallery set}. 
The targeted DOV approaches cover advanced algorithms across all categories: (1) backdoor watermarks, including BadNets \cite{gu2017identifying}, UBW \cite{li2022untargeted}, Label-Consistent \cite{turner2019label}, and Narcissus \cite{zeng2023narcissus}, encompassing clean-label, invisible, and untargeted poisoning techniques; (2) non-poisoning watermarks, including Radioactive Data \cite{sablayrolles2020radioactive}, ANW \cite{zou2022anti}, Domain Watermark \cite{guo2024domain}, Isotope \cite{wenger2024data}, and ML Auditor \cite{huang2024general}; and (3) fingerprints, including Dataset Inference \cite{maini2021dataset} and MeFA \cite{liu2022your}.

Table \ref{tab:escaping_advanced} presents the generalization and verification metrics for the models before and after evasion attacks (the teacher and surrogate student in Escaping DOV). All DOV methods successfully identify the \textit{vanilla} teacher trained directly on the copyright dataset. However, none meet their verification thresholds (\( \text{VSR} > 30\% \) or \( \text{p-value} < 0.01 \)) on the student produced by Escaping DOV, where VSR is near random guesses and p-values exceed 0.1 in most cases. Moreover, Escaping DOV incurs minimal impact on generalization: test accuracy on CIFAR-10 decreases by less than 1\%, while on the challenging Tiny ImageNet set, with no overlapping gallery classes, accuracy drops by only 2-3\%. Thus, Escaping DOV achieves a robust balance between universal evasion and preserved generalization.

\begin{figure}[htb]  
    \centering  
    \includegraphics[width=\linewidth]{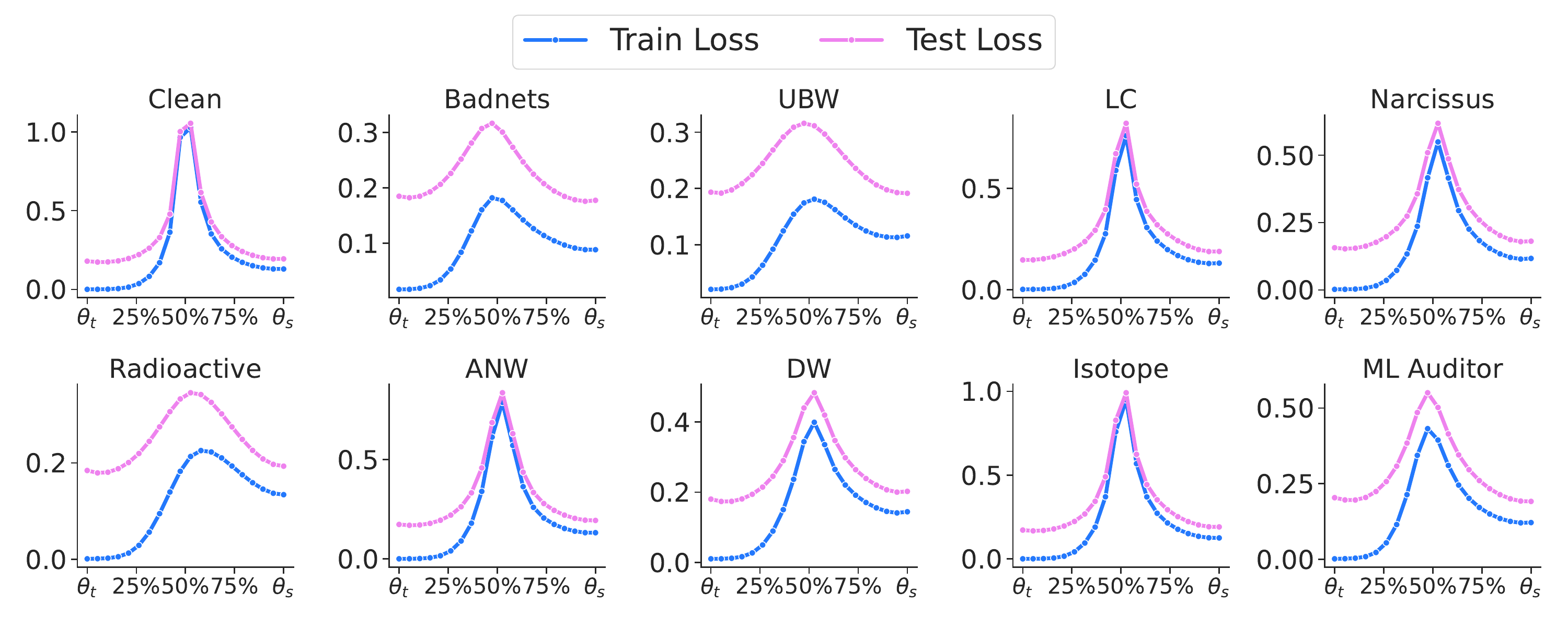}  
    \caption{Loss Barrier between Teacher and Student Parameters.}  
    \label{fig:mechanism}  
\end{figure}

We further observe that Escaping DOV encourages the student to diverge from the teacher’s predictive behavior, forming its own mechanism. In Figure \ref{fig:mechanism}, we linearly interpolate between teacher and student parameters, computing training and test losses at each point. The results reveal two key findings:
(1) A loss barrier appears during interpolation from teacher \( \theta_t \) to student \( \theta_s \). This lack of linear connectivity suggests that the surrogate student develops a distinct prediction \textit{mechanism}, bypassing the spurious features central to the teacher’s verification behavior \cite{lubana2023mechanistic}.
(2) For the \textit{teacher}, test loss is over \( 10^3 \) times the training loss, allowing fingerprints to easily infer dataset association. In contrast, the \textit{student’s} test loss is typically under twice the training loss, falling within the variance across subsets and making it far less distinguishable. Thus, Escaping DOV significantly suppresses both watermarks and fingerprints.

\subsection{Comparison with SOTA Evasion Attacks}
We compare with SOTA evasion attacks on CIFAR-10 across the seminal Badnets watermark and three most resistant DOVs from each category—Narcissus, Isotope, and Dataset Inference (see Table \ref{tab:escaping_advanced}). 
Results for all DOV methods are provided in Appendix \textcolor{iccvblue}{C.2}.
Evasion attacks include Fine-Pruning \cite{liu2018fine}, stronger combination of fine-tuning and pruning widely used in DOV evaluations, along with SOTA backdoor defenses across four stages: data sanitization with Meta-Sift \cite{287222}, robust training with Differential Privacy \cite{bu2023automatic}, backdoor unlearning with I-BAU \cite{zeng2022adversarial}, and input purification with ZIP \cite{shi2023black}. 
We also compare with methods that use knowledge distillation to mitigate backdoors, such as NAD \cite{li2021neural}, BCU \cite{pang2023backdoor} and ABD \cite{hong2023revisiting}, as well as IPRemoval \cite{zong2024ipremover}, which aims to eliminate model watermarks. Results in Table \ref{tab:comparison} and Appendix \textcolor{iccvblue}{C.2} show that Escaping DOV consistently surpasses all comparison methods in both generalization and evasion, and is the only approach that fully evades all DOV methods. For instance, only ABD and Escaping DOV reduce VSR below 40\% on the clean-label and invisible Narcissus watermark; however, ABD fails to evade Isotope, while Escaping DOV achieves nearly 10\% higher accuracy with superior evasion performance. 
We provide a rigorous analysis in Appendix \textcolor{iccvblue}{D}, explaining \textit{why SOTA evasion attacks, particularly those relying on distillation, are less effective than our Escaping DOV}.

\subsection{Ablations and Discussions}

\subsubsection{Transfer Set Curation}
\begin{figure}[htb]
    \centering
    \captionsetup[subfigure]{font=scriptsize}
    \captionsetup[subfigure]{skip=0pt}
    \begin{subfigure}{0.5\columnwidth}
        \includegraphics[width=\linewidth]{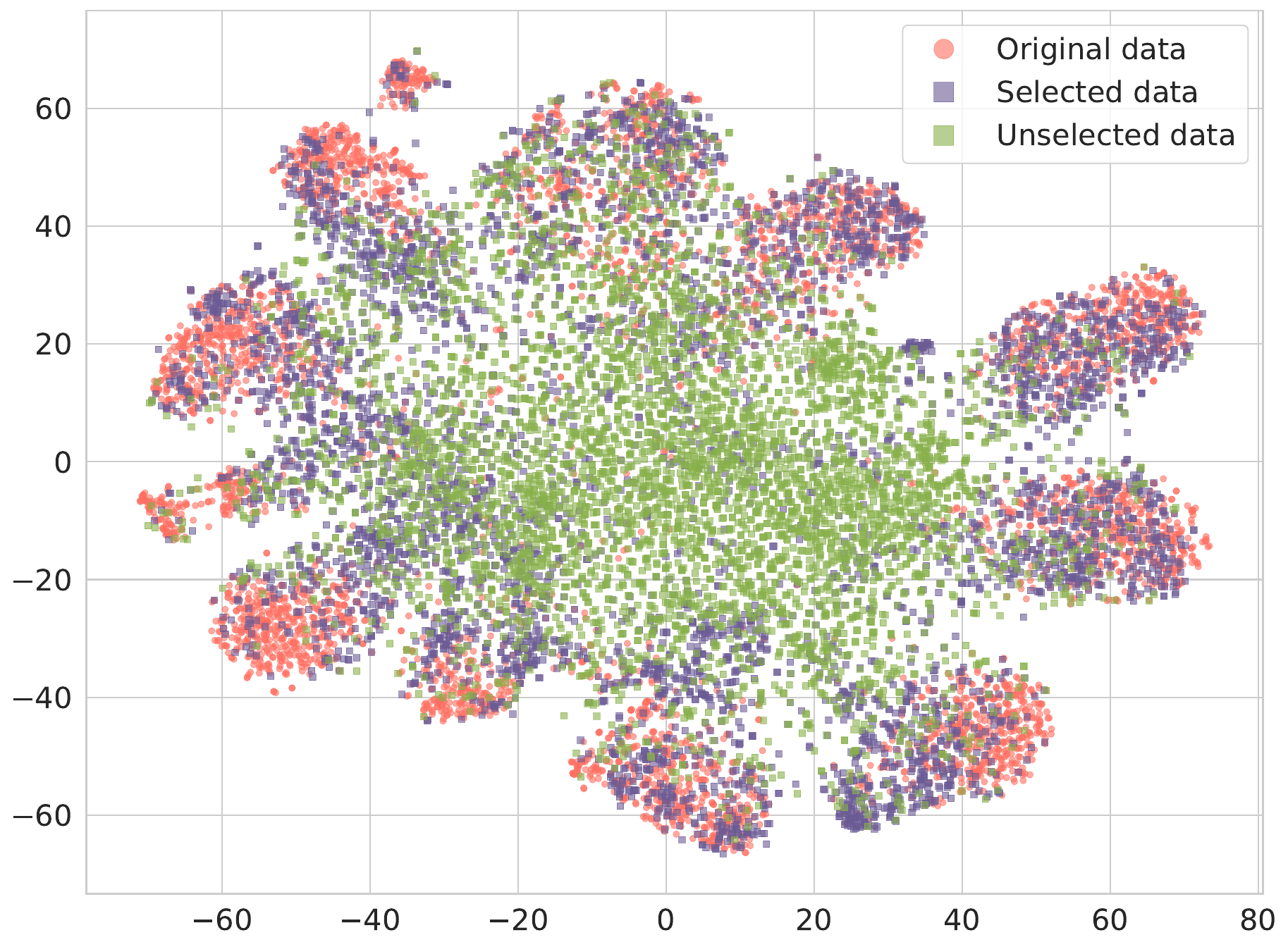} 
        \caption{Clean} 
        \label{fig:feature_space_a}
    \end{subfigure}%
    \hfill 
    \begin{subfigure}{0.5\columnwidth}
        \centering
        \includegraphics[width=\linewidth]{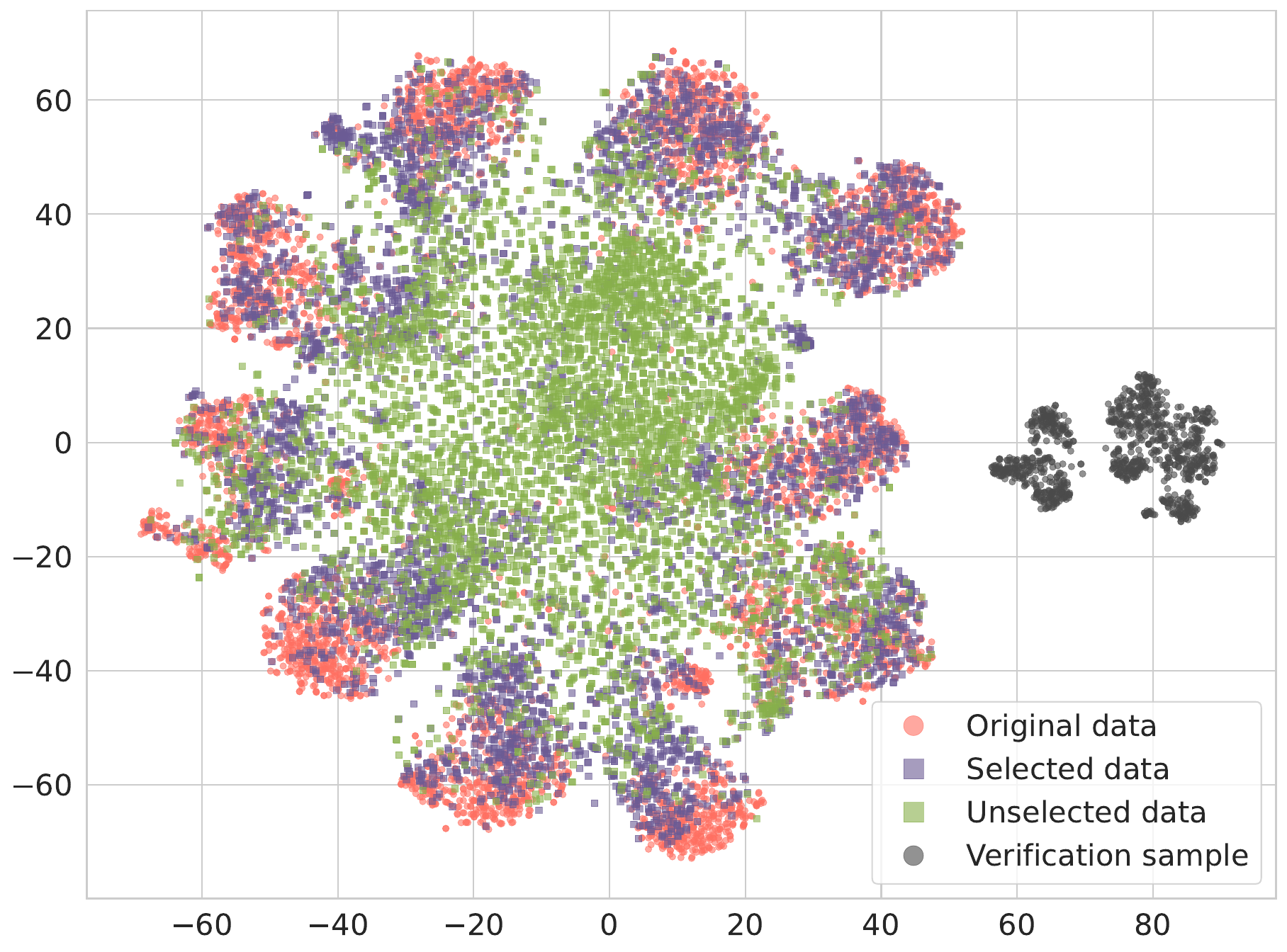} 
        \caption{Badnets} 
        \label{fig:feature_space_b}
    \end{subfigure}
    \caption{Sample Distribution in the Feature Space.} 
    \label{fig:feature_space}
\end{figure}

We begin with a \textit{qualitative analysis} of the transfer set. Figure \ref{fig:feature_space_a} visualizes the feature space for CIFAR-10 samples as well as selected and unselected ones from the transfer set, on a clean ResNet-18. The selected samples are notably closer to original CIFAR-10 samples. Figure \ref{fig:feature_space_b} illustrates the feature space of a Badnets-marked model, where trigger samples form distinct clusters with almost no transfer set samples nearby. The \textit{exclusive} and \textit{subtle} verification behaviors are rarely activated by OOD samples, and the curation process effectively excludes suspicious samples.

\textit{Quantitative analysis} follows, using Optimal Transport Dataset Distance (OTDD) \cite{alvarez2020geometric} to assess the distance from the target dataset to each proxy distribution, as shown in Table \ref{tab:ootd}. The curated transfer set is significantly closer to the target distribution than random sampling. Further, we compare the curated transfer set with random and unrelative samples on CIFAR-10 with Narcissus. Table \ref{tab:transfer_set} shows that transfer set curation not only improves generalization but also suppresses DOV by selecting more reliable samples.

\begin{table}[htb]
  \centering
  \caption{Optimal Transport Distances Between Datasets.}
  \resizebox{\linewidth}{!}{
    \begin{tabular}{c|cccc}
    \toprule[1.5pt]
    Distance & self  & LSVRC(\textbf{curation}) & LSVRC(random) & GTSRB \cite{stallkamp2011german} \\
    \midrule
    CIFAR-10  & 0.771  & 0.937  & 1.116  & 1.129  \\
    Tiny-Imagenet & 0.855 & 0.988 & 0.993 & 1.066\\
    \bottomrule[1.5pt]
    \end{tabular}%
    }
  \label{tab:ootd}
\end{table}%

\begin{table}[htb]
  \centering
  \caption{Escaping DOV across different Transfer Sets.}
  \resizebox{\linewidth}{!}{
    \begin{tabular}{c|ccc}
    \toprule[1.5pt]
    Transfer Set & GTSRB \cite{stallkamp2011german}  & LSVRC(random) & LSVRC(\textbf{curation}) \\
    \midrule
    \cellcolor[rgb]{.906, .902, .902} ACC(↑) & 66.80 ± 0.41 & 91.63 ± 0.49 & \textbf{94.37 ± 0.18} \\
    \cellcolor[rgb]{.906, .902, .902} VSR(↓) & 8.96 ± 1.63 & 7.87 ± 0.52 & \textbf{4.59 ± 0.64} \\
    \bottomrule[1.5pt]
    \end{tabular}%
    }
  \label{tab:transfer_set}
\end{table}%
\subsubsection{Selective Knowledge Transfer}
\begin{figure}[htb]  
    \centering  
    \includegraphics[width=\linewidth]{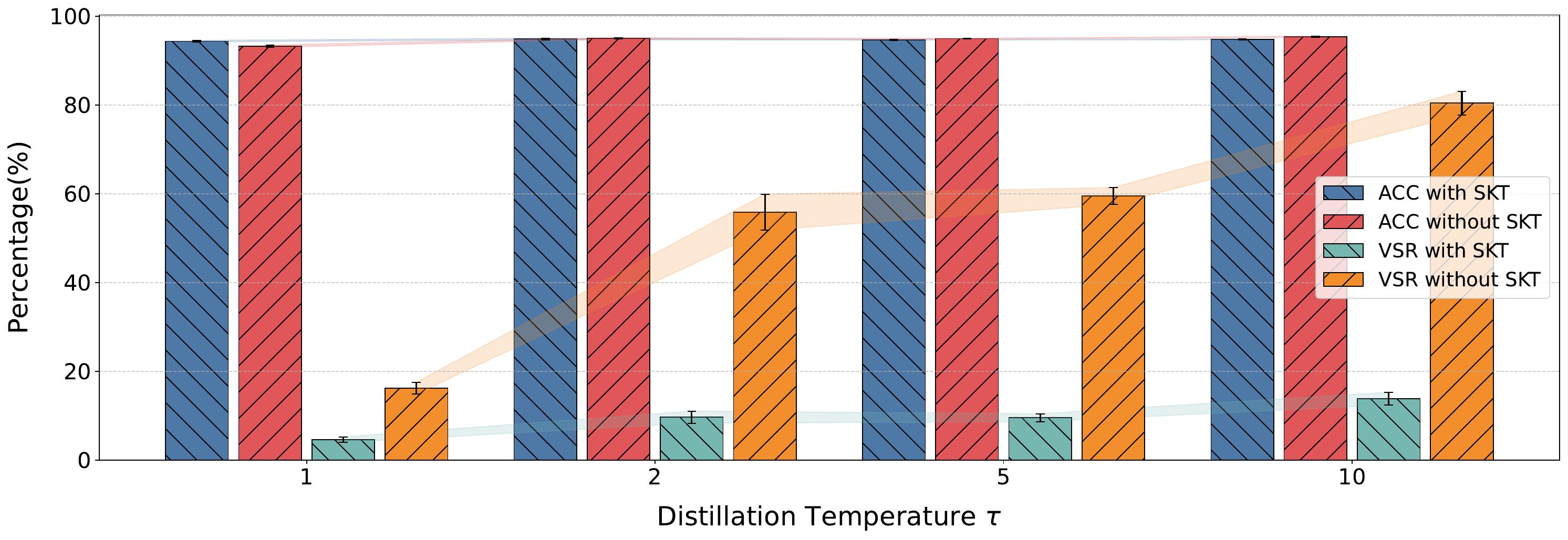}  
    \caption{Impact of SKT and Distillation Temperature.}  
    \label{fig:skt_temperature}  
\end{figure}

Figure \ref{fig:skt_temperature} illustrates the impact of Selective Knowledge Transfer (SKT) on the most resistant Narcissus watermark across distillation temperatures \(\tau\). As \(\tau\) increases, omitting SKT causes a substantial increase in VSR, while SKT consistently suppresses verification behavior across all temperatures. Additionally, VSR escalates sharply at higher temperatures, while test accuracy remains nearly unchanged. We hypothesis that task and verification knowledge possess distinct "boiling points", naturally separating at lower temperatures. Thus, we adopt \(\tau = 1\) as a reliable default.

We further compare the offline perturbation pool in SKT with direct UAT on the student. While UAT lowers VSR to 2.07\%, it incurs a 3.54\% accuracy reduction and more than triples computational cost. As SKT balances evasion and generalization, it serves as a lightweight approximation. Detailed time complexity is analysed in Appendix \textcolor{iccvblue}{C.8}.

\subsubsection{Escaping Adaptive Defenses}
We assume a \textit{knowledgeable} defender employing adaptive verification, specifically a semantic backdoor \cite{sun2024neural} and an anti-distillation backdoor \cite{ge2021anti}.
For the semantic backdoor, we assume the defender knows the gallery set and selects 500 "tench" samples from LSVRC-2012, relabeling them as "plane" and inserting them into CIFAR-10 to test their classification probability as "plane" (VSR). Since the teacher memorizes this verification behavior, using it for sample selection allows triggers enter the transfer set. For instance, DFND \cite{ge2021anti}, an entropy-based selection method, produces a student with VSR of 70.2\% by including 254 "tench" samples in the transfer set. In contrast, our curated transfer set contains fewer than 10 "tench" samples, reducing VSR to only 17.4\%. Even if the defender knows the gallery set, Transfer Set Curation still selects reliable samples. We also evaluate the entropy of teacher annotations: DFND's entropy is 2.90, while ours reaches 3.32, near the theoretical optimum $\log_2(10)$, indicating better class balance.

\begin{table}[htbp]
  \centering
  \caption{Escaping DOV against the Adaptive ADB Watermark.}
  \resizebox{\linewidth}{!}{
  \begin{tabular}{c|cccccc}
    \toprule[1.5pt]
    \multirow{2}[0]{*}{$\mu$} & \multicolumn{2}{c}{\bfseries Vanilla} & \multicolumn{2}{c}{\bfseries Evasion} & \multicolumn{2}{c}{\bfseries Independent} \\
    \cmidrule(r){2-3} \cmidrule(lr){4-5} \cmidrule(l){6-7}
           & \cellcolor[rgb]{.906, .902, .902} ACC & \cellcolor[rgb]{.906, .902, .902}VSR & \cellcolor[rgb]{.906, .902, .902}ACC & \cellcolor[rgb]{.906, .902, .902}VSR & \cellcolor[rgb]{.906, .902, .902}ACC & \cellcolor[rgb]{.906, .902, .902}VSR \\
    \midrule
    0.1   & 91.59\% &  100.00\% & 90.14\% & 94.42\% & 95.13\% & 92.89\% \\
     1     & 90.90\% & 100.00\% & 89.23\% & 78.48\% & 95.13\% & 71.74\% \\
    10    & 91.26\% & 12.88\% & 88.66\% & 14.21\% & 95.13\% & 10.01\% \\
    \bottomrule[1.5pt]
  \end{tabular}%
  }
  \label{tab:adaptive}%
\end{table}%

The anti-distillation backdoor (ADB) \cite{ge2021anti} is designed to transfer to student during distillation. As we employ distillation for knowledge transfer, we test ADB under a \textbf{hypothetical scenario} where \textbf{the defender controls the teacher’s training process to optimize the trigger}. Table \ref{tab:adaptive} reports the VSR across teacher, student, and unmarked models. A smaller $\mu$ allows larger perturbations, turning the trigger into an adversarial perturbation that induces \textbf{false positives} on unmarked models. Despite the defender's \textbf{unrealistic} control over the teacher, students' VSR remains closer to that of an unmarked model. As $\mu$ increases, ADB becomes ineffective. Thus, even under defender-controlled teacher training, Escaping DOV remains a potent threat.

\subsubsection{Enhanced Evasion Tactics}
Escaping DOV involves only post-training adjustments and operates orthogonally to pre-training, in-training and test-time strategies, all of which can collectively enhance evasion. As shown in Table \ref{tab:mix}, replacing data augmentation with mixup \cite{zhang2018mixup} during teacher training—an approach that mitigates backdoors and overfitting—further boosts evasion efficacy in conjunction with Escaping DOV on CIFAR-10. 

\begin{table}[htbp]
  \centering
  \caption{Mixup Teacher Training for Escaping DOV.}
  \resizebox{\linewidth}{!}{
  \begin{tabular}{c|cccc}
    \toprule[1.5pt]
         \bfseries DOV & \multicolumn{2}{c}{\bfseries Vanilla} & \multicolumn{2}{c}{\bfseries Evasion} \\
    \midrule
    \rowcolor[rgb]{.906, .902, .902} \bfseries Poisoning DOV &  ACC &  VSR &  ACC(↑) &  VSR(↓) \\
    \midrule
    \bfseries Narcissus & 94.04\% & 72.47\% & 93.03\% & 3.66\% \\
    \midrule
    \rowcolor[rgb]{.906, .902, .902} \bfseries Non-poisoning DOV &  ACC &  {p}-value &  ACC(↑) &  p-value(↑) \\
    \midrule
    \bfseries Isotope & 94.03\% & 0.2964 & 92.71\% & 0.7013 \\
    \bfseries Dataset Inference & 94.60\% & 0.0937 & 92.78\% & 0.5196 \\
    \bottomrule[1.5pt]
  \end{tabular}%
  }
  \label{tab:mix}%
\end{table}%

\subsubsection{Case Study on Domain-Specific Datasets}
\label{sec:domain-specific}
Datasets in vertical industries often require stronger protection than natural images due to collection challenges and privacy concerns. We conduct a case study on RAFDB (facial emotion) \cite{li2017reliable} and OrganCMNIST (medical imaging) \cite{yang2023medmnist} in Table \ref{tab:domain_specific}, with detailed results in Appendix \textcolor{iccvblue}{C.3}. Escaping DOV maintains generalization and efficacy despite unique distribution, while a larger domain-related gallery and VLM could further improve performance \cite{li2024flip, lu2024visual}.

\begin{table}[htb]
  \centering
  \caption{Escaping DOV on Domain-specific Datasets.}
  \resizebox{\linewidth}{!}{
    \begin{tabular}{cc|cccc}
     \toprule[1.5pt]
    \multicolumn{2}{c|}{\textbf{Settings}} & \multicolumn{2}{c}{\textbf{Vanilla}} & \multicolumn{2}{c}{\textbf{Evasion}} \\
    \midrule
     \rowcolor[rgb]{.906, .902, .902} \textbf{DOV}& \textbf{DataSet}      & ACC   & VSR   & ACC(↑) & VSR(↓) \\
     \midrule
          \multirow{2}[0]{*}{\textbf{Badnets}}& RAFDB & 85.21 ± 0.33 & 99.95 ± 0.02 & 80.53 ± 0.42&2.19 ± 0.06
  \\
          & OrganC & 95.32 ± 0.06 & 99.99 ± 0.01 & 90.57 ± 0.70 & 0.83 ± 0.15 \\
    \midrule
     \rowcolor[rgb]{.906, .902, .902} \textbf{DOV}&  \textbf{DataSet}     & ACC   & p-value & ACC(↑) & p-value(↑) \\
     \midrule
          \multirow{2}[0]{*}{\textbf{Isotope}}& RAFDB & 82.46 ± 0.59 & 1.50e-06 & 80.37 ± 0.13 & 1.05e-01 \\
          & OrganC & 94.63 ± 0.19 & 2.65e-03 & 88.96 ± 0.56 & 5.04e-01 \\
          \bottomrule[1.5pt]
    \end{tabular}%
    }
  \label{tab:domain_specific}
\end{table}%

\section{Conclusion}
In this paper, we reveal that Data Ownership Verification, widely used to prevent unauthorized model training, is vulnerable to strong evasion attacks. We introduce Escaping DOV, which leverages transfer set curation and selective knowledge transfer to achieve superior generalization and universal evasion capabilities. 
We propose that Escaping DOV serve as a evaluation framework to help establish truly reliable DOV methods, with plans for extensive cross-modal experiments to explore broader societal benefits.
\section*{Acknowledgments}
The work was supported by the National Natural Science Foundation of China under Grant 62271307 and 61771310.
{
    \small
    \bibliographystyle{ieeenat_fullname}
    \bibliography{main}
}

\newpage 
\clearpage
\maketitlesupplementary
\appendix
\renewcommand{\thesection}{\Alph{section}}
\startcontents 
{\centering\section*{Contents}\par}
\printcontents{}{1}{} 

\section*{Organization of the Supplementary Material}

This section provides an overview of the supplementary material. The core implementation of the Escaping DOV framework can be accessed via the \textbf{Anonymous Repository}: \url{https://github.com/dbsxfz/EscapingDOV}.

The supplementary material is organized as follows:
\begin{itemize}
\item \textbf{Section \textcolor{red}{A}}: A detailed overview of the Escaping DOV framework, including illustrative examples, implementation pipelines, and pseudo-code.
\item \textbf{Section \textcolor{red}{B}}: Descriptions and parameter configurations for DOV and baseline evasion methods.
\item \textbf{Section \textcolor{red}{C}}: Additional experimental results and visualizations, covering: (1) weaker evasion attacks used in previous DOV literature, (2) results of SOTA evasion attacks on all DOV methods, (3) Escaping DOV on copyright datasets with significant distribution shifts from the gallery set, (4) evaluations across model architectures, (5) the side benefits of Escaping DOV, (6) time complexity analyses, and (7) examples of transfer set curation.
\item \textbf{Section \textcolor{red}{D}}:
A rigorous analysis of why Escaping DOV successfully evades all DOV methods and why other SOTA evasion attacks fail against certain DOV approaches, particularly distillation-based ones.
\item \textbf{Section \textcolor{red}{E}}:
A discussion on DOV methods in other modalities and tasks, such as large language models (LLMs), and potential countermeasures against Escaping DOV.
\end{itemize}

\section{Detailed Algorithms of Escaping DOV}
\subsection{Zero-shot Prompt Learning}
As outlined in Section \textcolor{iccvblue}{3.2} of the main text, relying solely on a text template containing class names (e.g., “a photo of {class name}”) is insufficient for vision-language models (VLMs) to effectively capture the foreground semantics necessary to distinguish between target task categories \cite{pratt2023does}. Moreover, few-shot adaptation on the copyright dataset $\mathcal{D}$ inevitably introduces verification behaviors into the VLM \cite{bai2024badclip}, which impairs reliable transfer set selection.

To address these limitations, we adopt zero-shot prompt learning \cite{pratt2023does} to adapt the VLM to the target task by leveraging unbiased world knowledge embedded in a large language model (LLM). Specifically, we generate tailored image-prompts as a description set $Desc_{c_i}$ for each image category $c_i$ using GPT-4o-mini \cite{hurst2024gpt}, enhancing the performance of VLM.

\begin{figure}[htbp]  
    \centering  
    \includegraphics[width=\linewidth]{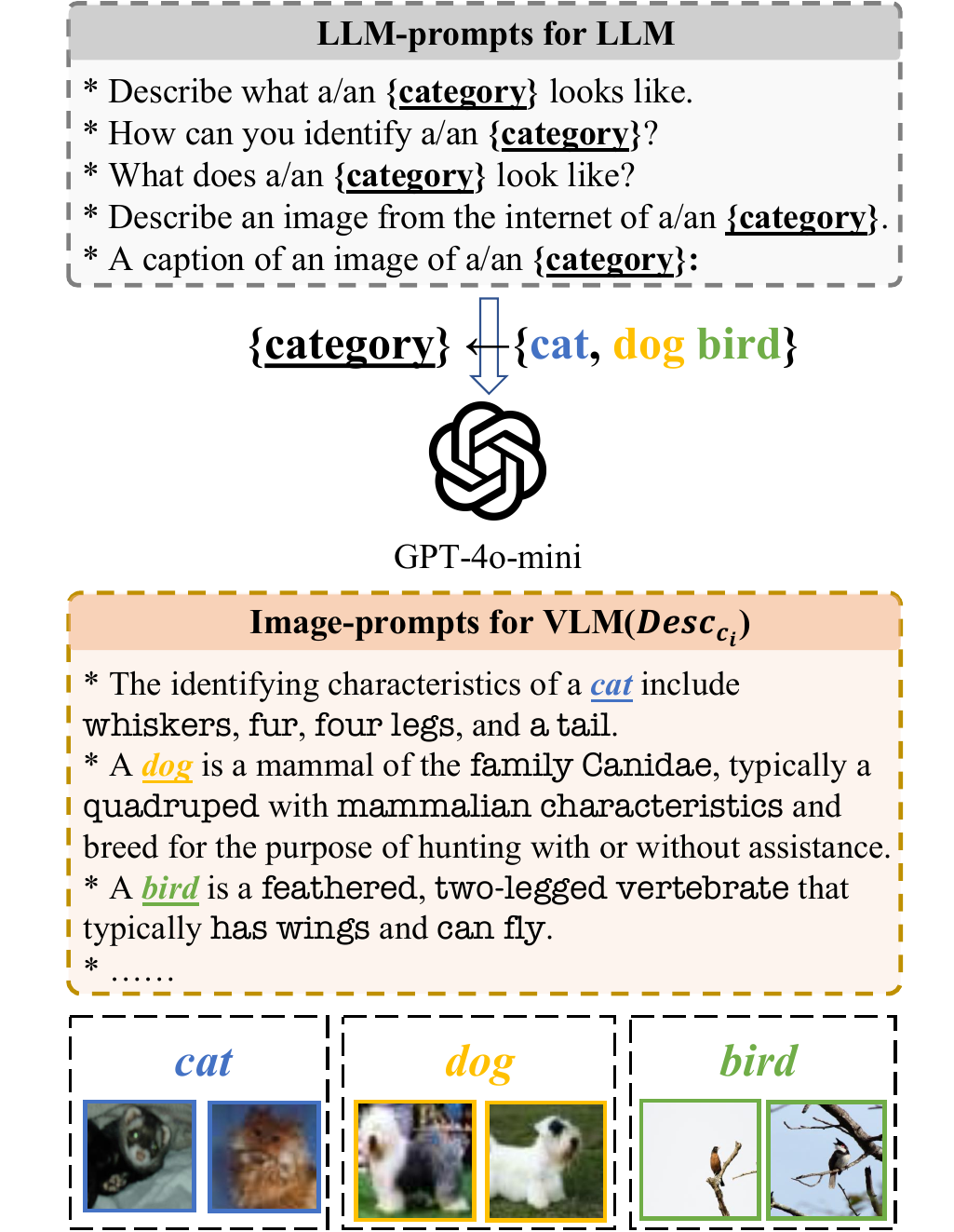}  
    \caption{Pipeline of Generating Image-prompts for VLM.}  
    \label{fig:promptGeneration}  
\end{figure}

The zero-shot prompt learning process is simplified into two steps, as illustrated in Figure \ref{fig:promptGeneration}. The distinctions between LLM-generated prompts for the LLM and image-prompts for the VLM are further detailed in Table \ref{tab:templateDiff}.
\begin{table*}[htbp]
  \centering
  \caption{Difference Between \textit{\textbf{LLM-prompts for LLM}} and \textit{\textbf{Image-prompts for VLM}}.}
    \resizebox{\linewidth}{!}{
  \begin{tabular}{c|c|c}
    \toprule[1.5pt]
    Aspect & \textit{\textbf{LLM-prompts for LLM}} & \textit{\textbf{Image-prompts for VLM}}  \\
    \midrule
    \multirow{1}{*}{Purpose}  & Instruct the LLM on how to describe a category.  & Serve as input to the VLM for zero-shot classification.  \\
    \midrule
    \multirow{1}{*}{Content}  & Generalized templates with a placeholder for category names  & Descriptive and category-specific sentences   \\
    \midrule
    \multirow{1}{*}{Generator}  & Designed manually & Generated by the LLM based on the corresponding prompt templates.  \\
    \midrule
    \multirow{1}{*}{Customization Level}  & Uniform templates used for multiple categories.  & Specific to the category being described.  \\
    \midrule
    \multirow{3}{*}{Examples}  & "Describe what a/an \textbf{\{\underline{\textcolor{darkgray}{category}}\}} looks like" & "A \textcolor{blue}{\underline{cat}} has four legs, a tail, and fur."  \\
    & "How can you identify a/an \textbf{\{\textcolor{darkgray}{\underline{category}}\}} + ?" & "A \textcolor{forestgreen}{\underline{bird}} generally has wings and feathers, and can fly." \\
    & "What does a/an \textbf{\{\underline{\textcolor{darkgray}{category}}\}} look like?" & "A \textcolor{deepyellow}{\underline{dog}} is a four-legged mammal of the family Canidae." \\
    \midrule

  \end{tabular}%
  }
  \label{tab:templateDiff}%
\end{table*}%
\subsubsection{Pipeline of Generating Image-prompts for VLM}


\begin{enumerate}
    \item Constructing \textit{\textbf{LLM-prompts for LLM}}: Manually craft a set of class-name-only general text templates, referred to as LLM-prompts, which are designed to elicit descriptive information about a given category. 
    \item Generating \textit{\textbf{Image-prompts for VLM}} using LLM : Input each LLM-prompts, filled with the specific category name $c_i$, into the LLM(such as GPT-4o-mini in our paper). The LLM generates multiple descriptive image-prompts $Desc_{c_i}$ for VLM that provide detailed visual descriptions of the category $c_i$. 
\end{enumerate}

\subsubsection{Examples of Prompts}

\paragraph{LLM-prompts for LLM}

\begin{itemize}
    \item[*] "Describe what a/an \textbf{\{\underline{\textcolor{darkgray}{category}}\}} looks like."
    \item[*] "How can you identify a/an \textbf{\{\underline{\textcolor{darkgray}{category}}\}}?"
    \item[*] "What does a/an \textbf{\{\underline{\textcolor{darkgray}{category}}\}} look like?"
    \item[*] "Describe an image from the internet of a/an \textbf{\{\underline{\textcolor{darkgray}{category}}\}}."
    \item[*] "A caption of an image of a/an \textbf{\{\underline{\textcolor{darkgray}{category}}\}}:"
\end{itemize}

\paragraph{Image-prompts for VLM}

\begin{itemize}
\item  When \textbf{\{\underline{\textcolor{darkgray}{category}}\}} = \textbf{\textcolor{blue}{\underline{cat}}}, generated Image-prompts are:
\begin{itemize}
    \item[*] "A \textbf{\textcolor{blue}{\underline{cat}}} has four legs, a tail, and fur."
    \item[*] "A typical house\textbf{\textcolor{blue}{\underline{cat}}} is small and has four legs."
    \item[*] "The identifying characteristics of a \textbf{\textcolor{blue}{\underline{cat}}} include whiskers, fur, four legs, and a tail."
    \item[*] "A \textbf{\textcolor{blue}{\underline{cat}}} is a small carnivorous mammal."
    \item[*] "The most common domestic \textbf{\textcolor{blue}{\underline{cat}}} is the brown tabby."
    \item[*] ......
\end{itemize}

\item When \textbf{\{\underline{\textcolor{darkgray}{category}}\}} = \textbf{\textcolor{forestgreen}{\underline{bird}}}, generated Image-prompts are:

\begin{itemize}
    \item[*] "The \textbf{\textcolor{forestgreen}{\underline{bird}}} has a long neck, short legs, and a long, thin beak."
    \item[*] "A \textbf{\textcolor{forestgreen}{\underline{bird}}} is a feathered, two-legged vertebrate that typically has wings and can fly."
    \item[*] "\textbf{\textcolor{forestgreen}{\underline{Bird}}}s are a type of vertebrate animal, characterized by feathers, toothless beaked mouths, the laying of hard-shelled eggs, a high metabolic rate, a four-chambered heart, and a strong yet lightweight skeleton."
    \item[*] "Some identifying characteristics of a \textbf{\textcolor{forestgreen}{\underline{bird}}} are that they have wings, feathers, and a beak."
    \item[*] "Most \textbf{\textcolor{forestgreen}{\underline{bird}}}s have wings, feathers, and beaks."
    \item[*] ......
\end{itemize}

\item  When \textbf{\{\underline{\textcolor{darkgray}{category}}\}} = \textbf{\textcolor{deepyellow}{\underline{dog}}}, generated Image-prompts are:
\begin{itemize}

    \item[*] "\textbf{\textcolor{deepyellow}{\underline{Dog}}}s are playful, friendly, and loyal animals."
    \item[*] "A \textbf{\textcolor{deepyellow}{\underline{dog}}} is a mammal of the family Canidae, typically a quadruped with mammalian characteristics and breed for the purpose of hunting with or without assistance."
    \item[*] "Some identifying characteristics of a \textbf{\textcolor{deepyellow}{\underline{dog}}} are that they are a mammal, have four legs, a tail, and bark."
    \item[*] "Identifying characteristics of a \textbf{\textcolor{deepyellow}{\underline{dog}}} include four legs, a tail, and fur."
    \item[*] "Most \textbf{\textcolor{deepyellow}{\underline{dog}}}s have four legs, a tail, and fur."
    \item[*] ......

\end{itemize}
\item When \textbf{\{\underline{\textcolor{darkgray}{category}}\}} = ...
\end{itemize}

\subsection{Transfer Set Curation}

As illustrated in Figure \textcolor{iccvblue}{1} of the main text, the transfer set curation process comprises three primary steps: (1) The VLM assigns each gallery set sample in $\mathcal{G}$ to a class \( t \) from the copyright dataset $\mathcal{D}$, thereby partitioning the gallery set $\mathcal{G}$ into \( K \) bins corresponding to the target task. (2) Within each bin, gallery samples are sorted in ascending order based on their distances to the distribution digest \( \text{Cent}_t \). (3) From each bin, samples are selected sequentially from the front of the sorted list. Only those samples for which the teacher model \( f_{\theta_t} \) predicts the same class \( t \) are included, until the number of selected samples matches the class count \( |\mathcal{D}_t| \) in the copyright dataset $\mathcal{D}$.

This process ensures two key properties of the resulting transfer set $\mathcal{T}$: (1) For all samples in $\mathcal{T}$, the teacher model \( f_{\theta_t} \) and the VLM produce consistent class predictions. This consistency implies that the teacher's predictions align, at least partially, with the inherent semantics of the input images, as endorsed by the VLM. It also rules out predictions driven by specific verification behaviors (e.g., backdoor watermark outputs on trigger samples). (2) Building on the first property, samples are prioritized based on their proximity to the corresponding distribution digest. This ensures that the selected samples are more representative of the target task distribution in $\mathcal{D}$. Together, these two properties render the transfer set both \textit{reliable} and \textit{informative}, facilitating effective task-oriented yet identifier-invariant knowledge transfer from the teacher model \( f_{\theta_t} \) to the student model \( f_{\theta_s} \).

To formally describe the transfer set curation process, the corresponding algorithm is presented in Algorithm \ref{alg:transfer_set_generation}. Detailed implementation details can be found in the Anonymous GitHub repository.

\begin{algorithm}[htbp]
\caption{Transfer Set Curation}
\label{alg:transfer_set_generation}
\begin{algorithmic}
\REQUIRE Copyright dataset $\mathcal{D}$, gallery set $\mathcal{G}$, teacher model $f_{\theta_t}$, visual encoder $E_v(\cdot)$ and text encoder $E_t(\cdot)$ of the pre-trained VLM, number of $\mathcal{D}$'s classes $K$ \\
\ENSURE Transfer set $\mathcal{T}$

\STATE $\mathcal{T} \gets \emptyset$
\STATE $bins \gets \{\mathcal{B}_1, \mathcal{B}_2, \ldots, \mathcal{B}_K\} \gets \emptyset$
\FOR{each class $c_t \in \{c_0, \ldots, c_K\}$ from $\mathcal{D}$}
    \STATE \COMMENT{Calculate the density centroid for class $c_t$}
    \STATE $Cent_{c_t} \gets \frac{1}{|\mathcal{D}_{c_t}|} \sum_{i \in \mathcal{D}_{c_t}} E_v(i)$  
    \STATE \COMMENT{Generate descriptions for $c_t$}
    \STATE $Desc_{c_t} \gets \text{LLM description generation}(c_t)$
    \STATE $r_{c_t} \gets \frac{1}{|Desc_{c_t}|} \sum_{i \in Desc_{c_t}}  E_t(i)$
\ENDFOR
    \STATE \COMMENT{\textcolor{BrickRed}{Assign images to bins based on VLM prediction}}
    \FOR{$I$ in $\mathcal{G}$}
        \STATE $c_{\text{max}} \gets \text{argmax}_{c}(\text{sim}(E_v(I), r_{c_t}))$
        \STATE $\mathcal{B}_{c_{\text{max}}} \gets \mathcal{B}_{c_{\text{max}}} \cup \{I\}$
    \ENDFOR
\FOR{$\mathcal{B}_{c_t}$ in $bins$}
\STATE \COMMENT{\textcolor{BrickRed}{Sort images in each bin by visual similarity}}
\STATE $\text{key\_func}(I) \gets \text{sim}(E_t(I), Cent_{c_t})$
\STATE $\mathcal{B}_{c_t} \gets \text{sorted}(\mathcal{B}_{c_t}, \text{key} = \text{key\_func}, \text{descending})$
    \STATE \COMMENT{\textcolor{BrickRed}{Filter images in each bin}}
    \STATE $counter \gets 0$
    \FOR{$I$ in $\mathcal{B}_{c_t}$}
        \STATE $l_{\text{T}} \gets f_{\theta_t}(I)$
        \IF{$l_{\text{T}} = c_t$}
            \STATE $\mathcal{T} \gets \mathcal{T} \cup \{I\}$
            \STATE $counter \gets counter + 1$
        \ENDIF
        \IF{$counter = |\mathcal{D}_{c_t}|$}
            \STATE \textbf{break}
        \ENDIF
    \ENDFOR
\ENDFOR
\end{algorithmic}
\end{algorithm}

\subsection{Selective Knowledge Transfer}

To filter out suspicious verification knowledge that reflects inherent (e.g., fingerprints) or artificial (e.g., watermarks) biases in the copyright dataset $\mathcal{D}$, we propose a Selective Knowledge Transfer framework for extraction task-oriented yet identifier-invariant knowledge from the teacher model \( f_{\theta_t} \) to the student model \( f_{\theta_s} \).

\begin{algorithm}[htbp]
\caption{Generating Perturbations}
\label{alg:perb}
\begin{algorithmic}
\REQUIRE Teacher model $f_{\theta_t}$, copyright dataset $\mathcal{D}$, number of perturbations $N$, number of iterations $I$, learning rate $\alpha$, scaling factor $\eta$, regularization factor $\lambda$
\ENSURE Perturbation pool $\{\delta\}$

\STATE Initialize $\{\delta\} \gets \emptyset$

\FOR{$n = 1$ to $N$}
    \STATE $\delta_n \gets 0$
    \FOR{$i = 1$ to $I$}
    \FOR{each batch $(\mathbf{x}, \mathbf{y}) \in \mathcal{D}$}
        \STATE $\mathbf{\hat{y}} \gets \arg\max f_{\theta_t}(\mathbf{x})$
        \STATE $\mathbf{x}_{\text{pert}} \gets \mathrm{clip}(\mathbf{x} + \delta_n, 0, 1)$
        \STATE $\mathbf{z}_{\text{pert}} \gets f_{\theta_t}(\mathbf{x}_{\text{pert}})$
        \STATE $\mathcal{L} \gets \mathrm{CrossEntropy}(\mathbf{z}_{\text{pert}}, \mathbf{\hat{y}}) - \lambda \|\delta_n\|_2^2$
        \STATE $\delta_n \gets \delta_n \textcolor{BrickRed}{\, + \,} \alpha \cdot \nabla_{\delta_n} \mathcal{L}$
        \STATE
        $\delta_n \gets \delta_n \cdot \min\left(1, \frac{\eta}{\|\delta_n\|_2}\right)$
        
    \ENDFOR
    
    \STATE Project \(\delta_n\) onto the norm (\(L_0\), \(L_2\), \(L_\infty\)) that maximizes \(\mathcal{L}\)
    \ENDFOR

    \STATE $\{\delta\} \gets \{\delta\} \cup \{\delta_n\}$
\ENDFOR

\RETURN $\{\delta\}$
\end{algorithmic}
\end{algorithm}

\begin{algorithm}[htbp]
\caption{Generating Corruption Chain}
\label{alg:corr}
\begin{algorithmic}
\REQUIRE Teacher model $f_{\theta_t}$, copyright dataset $\mathcal{D}$, corruption functions $\{C_1, C_2, \dots, C_k\}$, number of epochs $N$, genetic algorithms NSGA2
\ENSURE Optimal corruption chain $\mathbf{s}^*$
\STATE Initialize population $p$ of corruption sequences $\{\mathbf{s}_i\}_{i=1}^n$
\FOR{epoch $n = 1$ to $N$}
    \STATE Sample batch $(\mathbf{x}, \mathbf{y}) \sim \mathcal{D}$
    \FOR{each sequence $\mathbf{s}_i$ in population $p$}    
        \STATE \COMMENT{Apply corruption sequence $\mathbf{s}_i$ to $\mathbf{x}$}
        \STATE $\mathbf{x}_{\text{corr}} \leftarrow C_{s_i[1]}(C_{s_i[2]}(\dots C_{s_i[m]}(\mathbf{x})))$
        \STATE $\mathcal{L}_i \leftarrow -\mathrm{CrossEntropy}(f_{\theta_t}(\mathbf{x}_{\text{corr}}), \mathbf{y})$
    \ENDFOR
    \STATE Update population $p$ using NSGA2 to maximize $\mathcal{L}$
\ENDFOR
\STATE Return the optimal corruption chain $\mathbf{s}^*$
\end{algorithmic}
\end{algorithm}

This process begins by generating worst-case perturbations, as described in Algorithm \ref{alg:perb}, and constructing corruption chains, detailed in Algorithm \ref{alg:corr}, which are designed to mislead the teacher model \( f_{\theta_t} \) on $\mathcal{D}$. These perturbations and corruption chains are then incorporated into the distillation process, encouraging the surrogate student model \( f_{\theta_s} \) to develop invariance to such misleading biases. The algorithm of the knowledge transfer process is provided in Algorithm \ref{alg:selective_knowledge_transfer}.

\vspace{1em}
\begin{algorithm}[htbp]
\caption{Selective Knowledge Transfer}
\label{alg:selective_knowledge_transfer}
\begin{algorithmic}
\REQUIRE Teacher model $f_{\theta_t}$, student model $f_{\theta_s}$, perturbation pool $\{\delta\}$, corruption chain $\mathbf{s}^*$, transfer set $\mathcal{T}$, learning rate $\alpha$, number of epochs $N$
\ENSURE Optimized student model parameters $\theta_s$
\FOR{$\text{epoch} = 1$ to $N$}
    \FOR{each batch $(\mathbf{x}, \mathbf{y}) \in \mathcal{T}$}
        \STATE Randomly select an Operation from \{Skip, Perturbation, Corruption\}
        \IF{Operation is Skip}
            \STATE $\mathbf{x}' \gets \mathbf{x}$
        \ELSIF{Operation is Perturbation}
            \STATE Randomly select a perturbation $\delta$ from $\{\delta\}$
            \STATE $\mathbf{x}' \gets \mathbf{x} + \delta$
        \ELSIF{Operation is Corruption}
            \STATE $\mathbf{x}' \gets \mathbf{x} \circ \mathbf{s}^*$
        \ENDIF
        \STATE $y_t \gets f_{\theta_t}(\mathbf{x})$ \COMMENT{Teacher model prediction}
        \STATE $y_s \gets f_{\theta_s}(\mathbf{x}')$ \COMMENT{Student model prediction}
        \STATE $\mathcal{L} \gets KL(y_t \parallel y_s)$ \COMMENT{Compute KL divergence}
        \STATE $\theta_s \gets \theta_s - \alpha \nabla_{\theta_s} \mathcal{L}$ \COMMENT{Update student parameter}
    \ENDFOR
\ENDFOR
\RETURN $\theta_s$
\end{algorithmic}
\end{algorithm}

\section{Details of DOV Methods and Evasion Attacks}

\subsection{DOV Settings}

This section describes the settings of DOV methods targeted by the proposed Escaping DOV framework. These methods are categorized into three groups: \textit{backdoor watermarks}, \textit{non-poisoning watermarks}, and \textit{dataset fingerprints}.

\subsubsection{Backdoor Watermarks}
\begin{enumerate}
\item \textbf{Badnets} \cite{gu2017identifying}: A classic poison-label backdoor watermark that uses a checkerboard-style trigger (size 3×3 for CIFAR-10 and 5×5 for Tiny-ImageNet). The trigger flips the label of affected samples to a target class, causing misclassification. The poison rate (percentage of samples with triggers in $\mathcal{D}$) is set to 10\%.

\item \textbf{UBW} \cite{li2022untargeted}: An \textit{untargeted} backdoor watermark, identical to Badnets in other settings, except that the labels of poisoned samples are randomized instead of being fixed to a single target class. This induces non-deterministic misclassification behavior, making it harder for backdoor defenses assuming a specific target class (e.g., Neural Cleanse \cite{wang2019neural}).

\item \textbf{Label-Consistent} \cite{turner2019label}: A \textit{clean-label} backdoor method that modifies only samples from the target class using adversarial perturbations to associate the trigger with the target class. The trigger, similar to that of Badnets, is placed in all four corners of the image. Poison rates are set to 10\% for CIFAR-10 and 50\% for Tiny-ImageNet to ensure verification on directly trained models.

\item \textbf{Narcissus} \cite{zeng2023narcissus}: A \textit{clean-label} and \textit{invisible} backdoor watermark with a trigger constrained to a $L_\infty$ norm of 16/255. Poison rates are 10\% for CIFAR-10 and 50\% for Tiny-ImageNet to ensure effective verification. The trigger is optimized on a surrogate model and acts partially as a universal adversarial perturbation, achieving slightly higher verification success rates (e.g., the VSR for the student model in Escaping DOV on Tiny-ImageNet is about 10\%, comparable to unmarked models).
\end{enumerate}

\subsubsection{Non-Poisoning Watermarks}
\begin{enumerate}
\item \textbf{Radioactive Data} \cite{sablayrolles2020radioactive}: Optimizes a perturbation (carrier vector) to align with the model weights trained on watermarked data. A T-test determines verification results using the lower loss on perturbed samples compared to natural samples. The perturbation is constrained to a 16/255 $L_\infty$ norm, and the poison rate is 10\%.

\item \textbf{ANW} \cite{zou2022anti}: Applies color transformations in hue space to induce lower loss on transformed samples in the watermarked model. The poison rate is 10\%.

\item \textbf{Domain Watermark} \cite{guo2024domain}: Uses a domain generator to transform selected samples into a domain that is hard to generalize. This induces the model trained on watermarked data to produce higher confidence on samples from this domain. A convolutional domain generator is used, and the poison rate is 10\%.

\item \textbf{Isotope} \cite{wenger2024data}: Blends external features (a fixed LSVRC-2012 image) with samples in $\mathcal{D}$ to induce high-confidence predictions on mixed verification samples. The blend ratio of original samples to external features is 0.9:0.1, and the poison rate is 10\%.

\item \textbf{ML Auditor} \cite{huang2024general} generates two sets of samples with opposing perturbations, optimizing them to be close in the input space but distant in the feature space of a surrogate model. One set is publicly released, while the other is kept private. Hypothesis testing is then performed by comparing the target model’s losses on these two sets. The poison rate is 10\%.
\end{enumerate}

\subsubsection{Dataset Fingerprints}
\begin{enumerate}
\item \textbf{Dataset Inference} \cite{maini2021dataset}: Uses black-box adversarial perturbations to measure the distance from a sample to the decision boundary. Samples from the training set have higher distances compared to external samples, and this property is verified via a T-test. The method uses 100 samples each from the training and testing sets, requiring a total of 60,000 queries to generate adversarial perturbations.

\item \textbf{MeFA} \cite{liu2022your}: Builds a meta-classifier for membership inference to distinguish training samples from external ones. Verification results are aggregated across 100 samples as per the original settings.
\end{enumerate}

\begin{figure}[htbp]  
    \centering  
    \includegraphics[width=\linewidth]{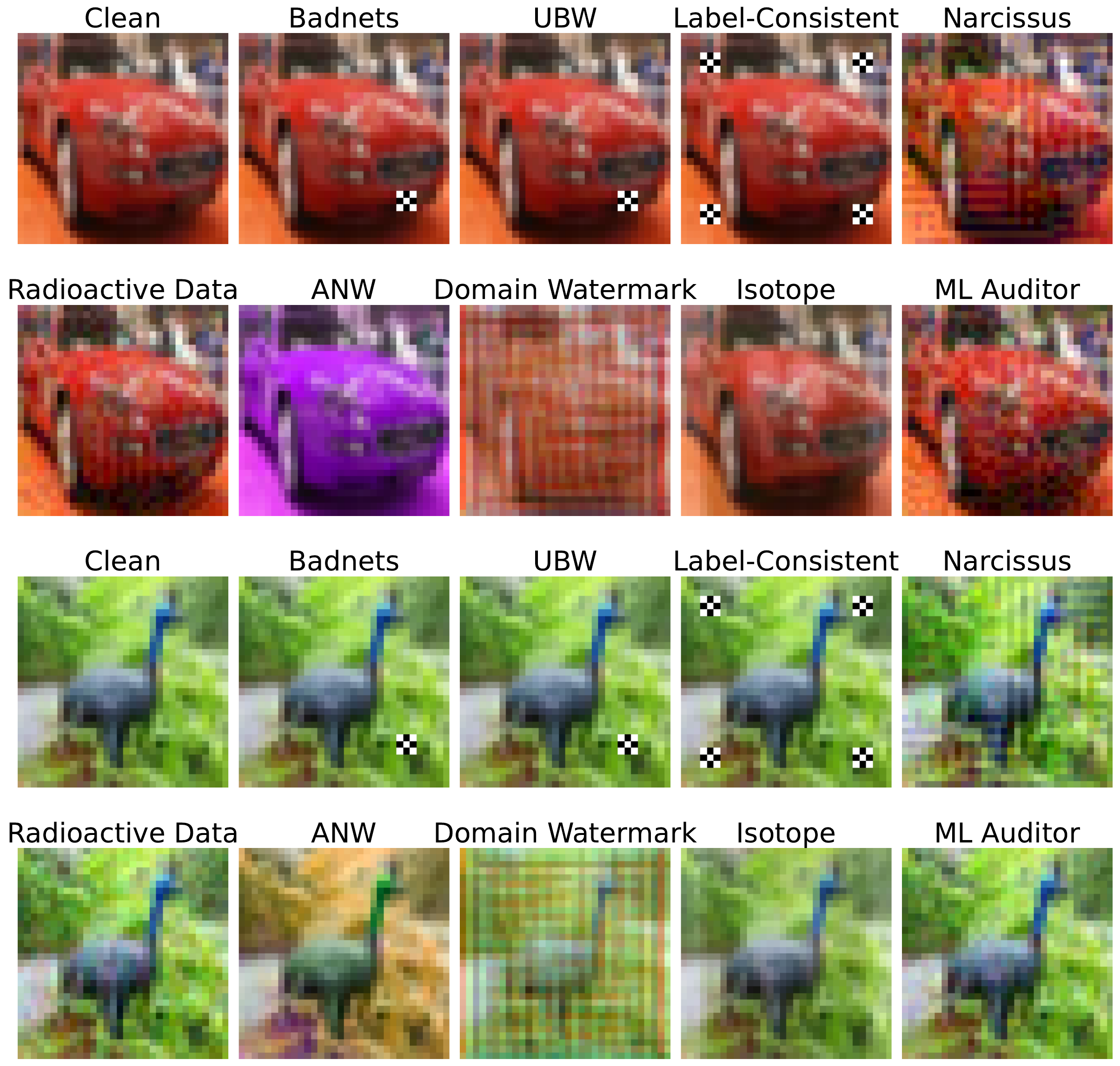}  
    \caption{Visual Samples of Different Watermarking Techniques.}  
    \label{fig:watermark_samples}  
\end{figure}

An example of watermarked samples for all methods is shown in Figure \ref{fig:watermark_samples}. It is evident that the trigger patterns are both \textit{exclusive} and \textit{subtle}, meaning that (1) they rarely appear in out-of-distribution (OOD) gallery sets and are therefore rarely activated by samples in the transfer set; (2) they do not alter the primary semantics of the images, while the verification behavior is orthogonal to the underlying distribution. This enables the surrogate student that avoids overfitting to specific biases to effectively mitigate such verification behavior.

\subsection{Evasion Attacks against DOV}

\begin{enumerate}
\item \textbf{Fine-pruning} \cite{liu2018fine} iteratively prunes and fine-tunes the trained model to remove embedded backdoors. In our experiments, 50\% of the parameters are pruned, and 5000 samples from LSVRC-2012 are used as the fine-tuning set.

\item \textbf{Meta-Sift} \cite{287222} sanitizes the training set $\mathcal{D}$ to create a smaller, clean subset for training. The clean set is set to be 50\% of the size of the original training set in our experiments.

\item \textbf{Differential Privacy} \cite{bu2023automatic} limits the contribution of individual samples to the model during training, thereby reducing the influence of poisoned samples. The privacy budget is set to $\epsilon = 2$.

\item \textbf{I-BAU} \cite{zeng2022adversarial} employs universal adversarial training with implicit gradients to unlearn backdoor triggers from a suspicious model. We use 5000 samples from LSVRC-2012 as the unlearning set and adhere to the settings in the original paper.

\item \textbf{ZIP} \cite{shi2023black} purifies test inputs using a pretrained diffusion model. We adopt guided diffusion \cite{dhariwal2021diffusion} for purification and follow the original paper's configuration.

\item \textbf{NAD} \cite{li2021neural} fine-tunes the original model as a teacher and \textit{initializes the student with parameters from the original model}. It removes backdoors by aligning intermediate features between the teacher and student using an attention pooling loss.

\item \textbf{BCU} \cite{pang2023backdoor} applies an adaptive layer-wise weight dropout strategy and prediction confidence screening during distillation to suppress backdoors. The dropout rates for each layer block are set to 0.1, 0.1, 0.2, 0.3, 0.5, and 0.5, respectively, and the confidence threshold is set to 0.9.

\item \textbf{ABD} \cite{hong2023revisiting} integrates backdoor detection \cite{cai2022randomized} with unlearning \cite{zeng2022adversarial} to suppress backdoors during distillation. We follow the detailed settings provided in the original paper.

\item \textbf{IPRemoval} \cite{zong2024ipremover} combines generative model inversion with data-free distillation and employs virtual ensemble distillation to remove model watermarks. We follow the original paper's settings in our implementation.
\end{enumerate}

\begin{table*}[htbp]
  \centering
  \caption{Weak Evasion Baselines Considered in Previous DOV Literature. VSR $>$ 30\% and p-value $<$ 0.01 Indicate \colorbox{lightred}{\textbf{Detection}}, Otherwise Successful \colorbox{lightgreen}{\textbf{Evasion}}.}
  \resizebox{\linewidth}{!}{
    \begin{tabular}{c|cccccccc}
    \toprule[1.5pt]
    \multirow{2}{*}{\textbf{Method}}& \multicolumn{2}{c}{\textbf{Badnets}} & \multicolumn{2}{c}{\textbf{Narcissus}} & \multicolumn{2}{c}{\textbf{Isotope}} & \multicolumn{2}{c}{\textbf{Dataset Inference}} \\
    \cmidrule(lr){2-3} \cmidrule(lr){4-5} \cmidrule(lr){6-7} \cmidrule(lr){8-9} 
    & ACC(↑) & VSR(↓) & ACC(↑) & VSR(↓) & ACC(↑) & p-value(↑) & ACC(↑) & p-value(↑) \\
    \midrule
    \textbf{AutoAugment} & 93.61  & \cellcolor{lightred}100.00  & 93.24  & \cellcolor{lightred}67.01  & 94.39  & \cellcolor{lightred}7.12e-03 & 94.53 & \cellcolor{lightred}1.69e-03 \\
    \textbf{Gaussian Blur} & 90.57  & \cellcolor{lightred}100.00  & 89.87  & \cellcolor{lightred}55.09  & 90.29  & \cellcolor{lightgreen}1.03e-01 & 90.28 & \cellcolor{lightred}7.60e-05 \\
    \textbf{Adv Training} & 84.99  & \cellcolor{lightred}100.00  & 85.15  & \cellcolor{lightgreen}10.04  & 84.89  & \cellcolor{lightgreen}2.18e-01 & 87.01 & \cellcolor{lightred}9.82e-05 \\
    \textbf{L2 Regularization} & 94.33  & \cellcolor{lightred}100.00  & 94.52  & \cellcolor{lightred}96.33  & 94.80  & \cellcolor{lightred}3.00e-05 & 95.76 & \cellcolor{lightred}7.18e-03 \\
    \textbf{Label Smoothing} & 93.80  & \cellcolor{lightred}100.00  & 94.33  & \cellcolor{lightred}93.86  & 94.33  & \cellcolor{lightred}7.03e-03 & 91.72 & \cellcolor{lightred}3.87e-03 \\
    \textbf{Distillation} & 94.09  & \cellcolor{lightred}100.00  & 93.65  & \cellcolor{lightred}78.01  & 94.40  & \cellcolor{lightred}9.73e-03 & 94.29 &\cellcolor{lightred}9.98e-04 \\
    \textbf{Zero-Shot Knowledge Distillation} & 90.46  & \cellcolor{lightred}60.74  & 91.56  & \cellcolor{lightred}52.13  & 93.59  & \cellcolor{lightred}8.33e-03 & 93.78 &\cellcolor{lightred}2.27e-03 \\
    \bottomrule[1.5pt]
    \end{tabular}%
    }
  \label{tab:weak}%
\end{table*}%

\section{Additional Experimental Results}

\subsection{Weaker Baseline Evasion Attacks Utilized in Previous DOV Research}

In this section, we present the performance of the weaker evasion attacks used in prior DOV research to evaluate the robustness of DOV methods, as shown in Table \ref{tab:weak}. \textbf{These weaker baseline evasion attacks prove entirely ineffective against the majority of DOV techniques, underscoring why the robustness of DOV methods has often been overestimated in previous studies}. Given that these attacks fail to induce meaningful evasion in most (if not all) DOV methods and are thus not directly comparable to our Escaping DOV attack, we report their results here rather than in the main text. A detailed analysis follows below:

Methods based on \textit{regularization} to prevent data provenance, such as the $L_{2}$ regularization in MeFA \cite{liu2022your}, label smoothing in ANW \cite{zou2022anti}, distillation as well as zero-shot knowledge distillation (ZSKD) in Dataset Inference \cite{maini2021dataset}, fail to evade any representative DOV techniques. While these methods occasionally reduce verification confidence, they can also amplify detection signals, as observed when $L_{2}$ regularization is applied to the Narcissus watermark. Since watermarks are carefully designed spurious features, and fingerprints are highly sensitive to confidence discrepancies between training and test data, these regularization methods are insufficient to achieve ideal generalization—i.e., preventing memorization of spurious features while maintaining nearly identical performance on training and test samples. Consequently, they fail to circumvent any DOV methods. 

Notably, both distillation and ZSKD, as considered in Dataset Inference \cite{maini2021dataset}, employ a knowledge transfer framework similar to our Escaping DOV. However, distillation uses the original copyright dataset as the transfer set, which essentially functions as a form of label smoothing. As a result, the mapping of watermark samples to their corresponding watermark labels is preserved, and the excessive memorization of copyright training data is entirely inherited by the student model, rendering this approach totally ineffective. ZSKD, on the other hand, employs generative adversarial training to invert training data from the teacher model, maximizing the prediction difference between teacher and student to synthesize data for distillation. However, this adversarial objective inadvertently reinforces the transfer of watermark behaviors and over-memorization—e.g., many watermark trigger patterns are even directly synthesized into the transfer set—aligning with observations in \cite{hong2023revisiting} and experimental results in \cite{zong2024ipremover}, as discussed in Section \ref{sec:d.2} of the Appendix. Consequently, ZSKD achieves only marginal evasion effects in the verification process.

Methods based on \textit{input perturbation}, which aim to \textit{disrupt watermark patterns} during training, include data augmentation (we use AutoAugment \cite{cubuk2019autoaugment}, a representative method that combines multiple data augmentation operations), adversarial training as considered in ANW \cite{zou2022anti}, and Gaussian blurring as explored in Isotope \cite{wenger2024data}. These methods occasionally succeed in bypassing certain DOV techniques, particularly invisible clean-label watermarks. This is because, for clean-label watermarks, the mapping between watermark triggers and watermark behaviors relies heavily on the precise presentation of the trigger pattern, while invisible watermarks generally lack sufficient resilience to input perturbations. However, these \textit{input perturbation} approaches are \textit{completely} ineffective against visible watermarks (e.g., BadNets) and fingerprints (e.g., DI and MeFA), rendering them unsuitable as a universal evasion baseline—i.e., an adversary with no prior knowledge of the specific DOV method employed by the copyright owner cannot reliably use them. Moreover, despite their occasional evasion success on certain DOV methods, adversarial training and Gaussian noise perturbations introduce significant performance degradation, further limiting their practical applicability.
\begin{table*}[htbp]
  \centering
  \caption{Evasion Attacks on other DOVs. VSR $>$ 30\% and p-value $<$ 0.01 Indicate \colorbox{lightred}{\textbf{Detection}}, Otherwise Successful \colorbox{lightgreen}{\textbf{Evasion}}.}
  \resizebox{\linewidth}{!}{
    \begin{tabular}{c|cccccccccccccc}
      \toprule[1.5pt]
      \multirow{2}{*}{\textbf{Method}} & \multicolumn{2}{c}{\textbf{UBW}} & \multicolumn{2}{c}{\textbf{Label-Consistent}} & \multicolumn{2}{c}{\textbf{Radioactive Data}} & \multicolumn{2}{c}{\textbf{ANW}} & \multicolumn{2}{c}{\textbf{Domain Watermark}} & \multicolumn{2}{c}{\textbf{ML Auditor}}& \multicolumn{2}{c}{\textbf{MeFA}} \\
      \cmidrule(lr){2-3} \cmidrule(lr){4-5} \cmidrule(lr){6-7} \cmidrule(lr){8-9} \cmidrule(lr){10-11} \cmidrule(lr){12-13} \cmidrule(lr){14-15}
          & ACC(↑)   & VSR(↓)   & ACC(↑)   & VSR(↓)   & ACC(↑)   & p-value(↑) & ACC(↑)   & p-value(↑) & ACC(↑)   & p-value(↑) & ACC(↑)   & p-value(↑) & ACC(↑)   & p-value(↑)\\
      \midrule
    \textbf{Fine-pruning} & 85.60 & \cellcolor{lightred}77.04 & 86.69 & \cellcolor{lightred}55.10 & 87.45 & \cellcolor{lightgreen}0.5303 & 86.83 & \cellcolor{lightgreen}0.8120 & 86.74 & \cellcolor{lightgreen}0.7299 &  85.72 & \cellcolor{lightred}1.9e-10 & 86.66 & \cellcolor{lightgreen}0.1164
    \\
    \textbf{Meta-Sift} & 86.85 & \cellcolor{lightgreen}12.17 & 87.03 & \cellcolor{lightred}82.28 & 87.35 & \cellcolor{lightgreen}0.5565 & 86.64 & \cellcolor{lightgreen}0.9665 & 88.48 & \cellcolor{lightgreen}0.9999 &83.61 & \cellcolor{lightgreen}0.6471& 88.70 &\cellcolor{lightred}0.0006  
    \\
    \textbf{Differential Privacy}    & 87.22 & \cellcolor{lightred}63.33 & 92.73 & \cellcolor{lightred}96.31 & 93.56 & \cellcolor{lightgreen}0.8248 & 91.15 & \cellcolor{lightred}0.0078 & 92.89 & \cellcolor{lightred}1.5e-23 & 93.06 & \cellcolor{lightred}6.2e-27 & 92.97 & \cellcolor{lightred}1.0e-17  
   \\
   \textbf{I-BAU} & 89.07 & \cellcolor{lightgreen}6.46  & 89.31 & \cellcolor{lightgreen}16.46 & 90.37 & \cellcolor{lightgreen}0.1484  & 92.68 & \cellcolor{lightgreen}0.3496  & 84.16 & \cellcolor{lightgreen}0.1129  & 90.98 & \cellcolor{lightred}4.9e-03 & 89.51 & \cellcolor{lightgreen}0.9993 
   \\
    \textbf{ZIP}   & 83.92 & \cellcolor{lightred}64.70 & 82.57 & \cellcolor{lightred}79.80 & 83.13 & \cellcolor{lightgreen}0.0129 & 83.04 & \cellcolor{lightgreen}0.1758 & 83.81 & \cellcolor{lightgreen}0.3698 & 78.79 & \cellcolor{lightred}7.4e-05 & 83.26 & \cellcolor{lightred}0.0004  
    \\
    \textbf{NAD}   & 86.76 & \cellcolor{lightred}91.83 & 92.23 & \cellcolor{lightred}62.29 & 92.16 &\cellcolor{lightgreen}0.0942 & 92.16 & \cellcolor{lightred}0.0014 & 90.27 &\cellcolor{lightred}8.8e-06 & 89.40 &\cellcolor{lightred}2.3e-05 & 91.56 &\cellcolor{lightred}0.0005 
    \\
    \textbf{BCU}   & 92.26 & \cellcolor{lightgreen}2.21  & 93.17 & \cellcolor{lightgreen}17.24 & 93.46 & \cellcolor{lightgreen}0.4798  & 92.91 & \cellcolor{lightgreen}0.8316  & 92.81 & \cellcolor{lightgreen}\textbf{1.0000}  & 91.63 & \cellcolor{lightgreen}0.4081 & 93.15 & \cellcolor{lightred}0.0014  
    \\
    \textbf{ABD}   & 82.62 & \cellcolor{lightgreen}5.28  & 82.92 & \cellcolor{lightgreen}11.41 & 88.23 & \cellcolor{lightgreen}0.2167  & 92.58 & \cellcolor{lightgreen}0.1300  & 83.36 & \cellcolor{lightgreen}0.9998 & 85.51 & \cellcolor{lightgreen}0.6230 & 88.26 & \cellcolor{lightgreen}0.9996 
    \\
    \textbf{IPRemoval} & 82.77 & \cellcolor{lightgreen}3.29  & 84.92 & \cellcolor{lightgreen}4.18  & 84.96 & \cellcolor{lightgreen}0.4234  & 84.14 & \cellcolor{lightgreen}\textbf{1.0000}  & 82.91 & \cellcolor{lightgreen}0.9999  &88.18 & \cellcolor{lightgreen}0.7517 & 85.47 & \cellcolor{lightgreen}0.9999
    \\
    \textbf{Escaping DOV (Ours)}  & \textbf{93.41} & \cellcolor{lightgreen}\textbf{1.74}  & \textbf{93.19} & \cellcolor{lightgreen}\textbf{3.74}  & \textbf{94.07} & \cellcolor{lightgreen}\textbf{0.9450} & \textbf{93.91} & \cellcolor{lightgreen}\textbf{1.0000} & \textbf{93.90} & \cellcolor{lightgreen}\textbf{1.0000} &\textbf{93.45} &\cellcolor{lightgreen}\textbf{0.7831} & \textbf{93.93} & \cellcolor{lightgreen}\textbf{1.0000}
    \\

      \bottomrule[1.5pt]
    \end{tabular}
  }
  \label{tab:moredov}
\end{table*}

\subsection{Additional Results of SOTA Evasion Attacks on Other DOV Methods}
In Section \textcolor{iccvblue}{4.3} of the main text, four representative DOV methods are selected. BadNets is a seminal work, while Narcissus, Isotope, and Dataset Inference are the most robust methods in their respective categories—backdoor watermarks, non-poisoning watermarks, and fingerprints—against our Escaping DOV attack. These methods exhibit the highest VSR or the lowest p-value after evasion in Table \textcolor{iccvblue}{1} of the main text, while the detection signal of the student models after evasion on other DOVs show no significant difference compared to independent models that have not used the copyright data.

We present the results of all SOTA evasion strategies on the remaining seven DOV methods in Table \ref{tab:moredov}. Despite selecting the most advanced and powerful backdoor mitigation and privacy-enhancing techniques for comparison, our Escaping DOV remains superior in both evasion and generalization performance. Notably, it is the only method that successfully escapes all 11 DOVs. In contrast, I-BAU, ABD, and IP-Removal fail on Narcissus and/or Isotope (as shown in Table \textcolor{iccvblue}{2} of the main text), with significantly weaker generalization performance.
Further in-depth analyses of why our Escaping DOV successfully evades all types of DOVs and why other SOTA evasion techniques fail against certain DOVs are provided in Section \ref{sec:d.1} and \ref{sec:d.2} of the Appendix.

\label{sec:c.2}
\begin{table*}[htbp]
  \centering
  \caption{Escaping DOV on Copyright Sets with Large Distribution Shift from the Gallery Set (\textcolor{BrickRed}{\textbf{ImageNet-1k}}). VSR $>$ 30\% and p-value $<$ 0.01 Indicate \colorbox{lightred}{\textbf{Detection}}, Otherwise Successful \colorbox{lightgreen}{\textbf{Evasion}}.}
  \resizebox{\linewidth}{!}{
    \begin{tabular}{cc|cccccccc}
    \toprule[1.5pt]
    \multicolumn{2}{c|}{\multirow{2}[0]{*}{\textbf{Copyright Set}}} & \multicolumn{2}{c}{\textbf{Badnets}} & \multicolumn{2}{c}{\textbf{Narcissus}} & \multicolumn{2}{c}{\textbf{Isotope}} & \multicolumn{2}{c}{\textbf{Dataset Inference}} \\
    \cmidrule(lr){3-4} \cmidrule(lr){5-6} \cmidrule(lr){7-8} \cmidrule(lr){9-10}
    \multicolumn{2}{c|}{} & ACC(↑) & VSR(↓) & ACC(↑) & VSR(↓) & ACC(↑) & p-value(↑) & ACC(↑) & p-value(↑) \\
    \midrule
    \multirow{2}[0]{*}{\textbf{GTRSB}} & Vanilla & 98.12 & \cellcolor{lightred}100.00 & 98.33 & \cellcolor{lightred}47.85 & 98.63 & \cellcolor{lightred}0.0040 & 98.68 & \cellcolor{lightred}0.0052  \\
          & Evasion & 95.23 & \cellcolor{lightgreen}3.20   & 94.84 & \cellcolor{lightgreen}9.76  & 95.67 & \cellcolor{lightgreen}0.4935 & 96.34 & \cellcolor{lightgreen}0.1130 \\
    \midrule
    \multirow{2}[0]{*}{\textbf{FER2013}} & Vanilla & 66.62 & \cellcolor{lightred}100.00 & 68.01 & \cellcolor{lightred}92.15 & 67.71 & \cellcolor{lightred}0.0006 & 68.12 & \cellcolor{lightred}4.90e-08 \\
          & Evasion & 62.19 & \cellcolor{lightgreen}7.51  & 62.94 & \cellcolor{lightgreen}12.43 & 63.31 & \cellcolor{lightgreen}0.1867 & 62.59 & \cellcolor{lightgreen}0.1429 \\ 
    \midrule
    \multirow{2}[0]{*}{\textbf{ImageNet-R}} & Vanilla & 65.33 & \cellcolor{lightred}100.00 & 66.90 & \cellcolor{lightred}57.29 & 66.23 & \cellcolor{lightred}0.0023 & 66.73 & \cellcolor{lightred}1.88e-18 \\
          & Evasion & 63.47 & \cellcolor{lightgreen}1.92  & 64.60 & \cellcolor{lightgreen}4.23  & 63.97 & \cellcolor{lightgreen}0.4032 & 63.83 & \cellcolor{lightgreen}0.1094 \\
    \midrule
    \multirow{2}[0]{*}{\textbf{ImageNet-Sketch}} & Vanilla & 83.30 & \cellcolor{lightred}81.24 & 85.66 & \cellcolor{lightred}53.09 & 86.25 & \cellcolor{lightred}0.0069 & 86.05 & \cellcolor{lightred}2.88e-06 \\
          & Evasion & 78.98 & \cellcolor{lightgreen}12.18 & 81.93 & \cellcolor{lightgreen}9.38  & 83.69 & \cellcolor{lightgreen}0.5192 & 82.91 & \cellcolor{lightgreen}0.0789 \\
    \bottomrule[1.5pt]
    \end{tabular}%
  }
  \label{tab:shift}%
\end{table*}

\subsection{Copyright and Gallery datasets with Large Distribution Shift}
\label{sec:c.3}
The primary experiments presented in the main text are conducted in scenarios where the copyright datasets are CIFAR-10 and Tiny-ImageNet, with the gallery set being LSVRC-2012 (ImageNet-1K). Alough ImageNet-1K contains images corresponding to some CIFAR-10 classes, it is not a superset of CIFAR-10. For instance, the "deer" class in CIFAR-10 lacks any direct counterpart in ImageNet-1K. \textbf{To establish a more challenging setting with no semantic overlap between the gallery set and the copyright set, we remove all 200 overlapping classes from ImageNet-1K (the gallery set) when conducting experiments on Tiny-ImageNet}. Even under this stringent condition, Escaping DOV continues to demonstrate strong performance. 

Since CIFAR-10, Tiny-ImageNet, and LSVRC-2012 are all natural image datasets, \textbf{we further investigate an even more challenging scenario where the copyright dataset consists of data from specific, hard-to-obtain vertical domains, while the gallery set remains the easily accessible natural images from LSVRC-2012}. 
In Table \textcolor{iccvblue}{7} of the main text, we present results for two copyright datasets with distributions that are entirely distinct from the gallery set (LSVRC-2012): the face recognition dataset RAFDB and the medical diagnosis dataset OrganCMNIST. In both cases, Escaping DOV maintains strong performance. Additionally, in Table \ref{tab:shift} of the Appendix, we extend our evaluation to four other copyright datasets that exhibit substantial distributional differences from the gallery set: the traffic sign dataset GTRSB \cite{stallkamp2011german}, the facial emotion dataset FER2013 \cite{goodfellow2013challenges}, and two datasets featuring artistic and non-photorealistic styles, ImageNet-R \cite{hendrycks2021many} and ImageNet-Sketch \cite{wang2019learning}. Given that ImageNet-Sketch contains only 50 samples per class, we sample 100 classes to ensure that the directly trained teacher model achieves meaningful generalization capacity.

Across all datasets, Escaping DOV evades detection successfully with minor degradation in generalization. Notably, a ResNet-18 trained on the original ImageNet-1k attains test accuracies of only 33.2\% and 39.69\% on ImageNet-R and ImageNet-Sketch, respectively, which is over 30\% lower than the performance of the student model obtained through Escaping DOV. This highlights the advantages of training on stolen target domain data and underscores the effectiveness of our knowledge transfer approach. We attribute this effectiveness to two key factors: (1) the adaptability of knowledge distillation to intermediary transfer sets \cite{frank2024makes}, and (2) the role of Transfer Set Curation (TSC) in Escaping DOV. For instance, while LSVRC-2012 does not contain a class directly corresponding to CIFAR-10's "deer" category, TSC effectively selects samples with highly similar semantic features as surrogates (see Figure \ref{fig:cifar10-2} and Section \ref{sec:c.7} in the Appendix).

Furthermore, these vertical-domain copyright datasets typically contain a limited number of samples per class. To ensure effective data provenance with the original (teacher) model, we apply a high watermark rate of 10\% (and 50\% for Narcissus). This results in the number of watermark samples often exceeding the number of clean samples in the target class. Consequently, the verification success rate (VSR) is slightly higher, and the p-value is slightly lower after our evasion attack compared with other datasets. In real-world scenarios where copyright owners are unable to watermark such a large volume of data, the evasion effectiveness of our Escaping DOV is expected to be even more pronounced.

\subsection{Why Choose OOD Data as the Transfer Set in Escaping DOV}
In addition to carefully selecting a transfer set from a large-scale OOD gallery using Transfer Set Curation (TSC), another possible approach for constructing the transfer set in Escaping DOV could involve using a small amount of clean IND data (if available), as demonstrated by baseline evasion attacks such as I-BAU \cite{zeng2022adversarial} and NAD \cite{li2021neural} do. However, we now analyze the advantages of curating a transfer set from a large-scale OOD gallery.

Firstly, due to privacy concerns and time-sensitive constraints, even a small amount of clean in-distribution (IND) data can be difficult to obtain, such as the latest medical images of new COVID-19 variants. Our Escaping DOV makes no assumptions regarding the availability of clean IND data (as demonstrated in Section \ref{sec:c.3}, where we show that a fixed natural image gallery set is applicable to any distribution of the copyright dataset). \textbf{In fact, while a small amount of clean IND data is insufficient to perform an evasion attack, it can certainly enhance the effectiveness of our Escaping DOV}.

We provide two examples to support this:
(1) We re-tested NAD \cite{li2021neural} on CIFAR-10 using 5\% clean IND samples. However, this did not lead to a significant improvement in performance, as NAD still failed on 6 out of 11 DOVs.
(2) Taking GTRSB as an example, when 5\% clean IND data was used in Escaping DOV, the student model achieved an accuracy of only 72.67\%, which is over 20\% lower than when the transfer set was curated using TSC from the OOD LSVRC-2012. However, when both OOD transfer set curation and the additional 5\% clean IND data were used together, the student accuracy further improved to 97.18\%.


\begin{table*}[htbp]
  \centering
  \caption{Escaping DOV Across Model Backbones on CIFAR-10.}
  \resizebox{\linewidth}{!}{
    \begin{tabular}{c|cccc|cccc|cccc}
      \toprule[1.5pt]
      \multirow{2}[0]{*}{\textbf{DOV}} &\multicolumn{4}{c|}{\textbf{ResNet-18}} & \multicolumn{4}{c|}{\textbf{Efficientnet v2}} & \multicolumn{4}{c}{\textbf{Swin Transformer v2}} \\
     \cmidrule(r){2-5} \cmidrule(l){6-9} \cmidrule(l){10-13}
      & \multicolumn{2}{c}{Vanilla} & \multicolumn{2}{c|}{Evasion} & \multicolumn{2}{c}{Vanilla} & \multicolumn{2}{c|}{Evasion}&\multicolumn{2}{c}{Vanilla} & \multicolumn{2}{c}{Evasion} \\
      \midrule
      \rowcolor[rgb]{.906, .902, .902} \textbf{Poisoning DOV}& ACC & VSR & ACC(↑) & VSR(↓) & ACC & VSR & ACC(↑) & VSR(↓) & ACC & VSR & ACC(↑) & VSR(↓) \\
      \midrule
     
    \textbf{Badnets} &94.44 &  100.00 & 93.46 & 1.36 & 94.36 & 100.00   & 92.51 & 1.07  & 90.57 & 100.00   & 89.78 & 1.56 \\
    \textbf{Narcissus} &94.76  & 87.34  & 94.37  & 4.59 & 94.85 & 89.93 & 93.58 & 1.29  & 91.36 & 92.71 & 89.98 & 2.47 \\

      \midrule
    \rowcolor[rgb]{.906, .902, .902} \textbf{Non-Poisoning DOV} & ACC  & p-value & ACC(↑) & p-value(↑) & ACC   & p-value & ACC(↑) & p-value(↑)& ACC   & p-value & ACC(↑) & p-value(↑) \\
    \midrule
\textbf{Isotope} &94.75 & 2.87e-03 & 93.99  & 2.84e-01& 95.2  & 1.11e-03 & 93.37 & 1.24e-01 & 91.17 & 7.61e-03 & 89.96 & 9.51e-02 \\
    \textbf{Dataset Inference} &94.83  & 1.87e-03 & 93.97  & 4.76e-01& 94.88 & 1.75e-05 & 93.35 & 1.47e-01 & 91.23 & 6.86e-03 & 90.07 & 1.23e-01 \\

    \bottomrule[1.5pt]
    \end{tabular}
  }
  \label{tab:escaping_backbone}
\end{table*}

\subsection{Escaping DOV with Advanced Backbones}

In the main text, we evaluate Escaping DOV using the relatively small ResNet-18 backbone. Here, we extend our experiments to larger and more advanced backbones, namely EfficientNet v2 \cite{tan2021efficientnetv2} and Swin Transformer v2 \cite{liu2022swin}. As shown in Table \ref{tab:escaping_backbone}, Escaping DOV continues to perform effectively, successfully evading all DOV methods on advanced model backbones without additional parameter tuning, despite that larger models, such as the Swin Transformer v2, exhibit severe overfitting on CIFAR-10.

\subsection{Robustness as a By-product}
\begin{table}[htbp]
  \centering
  \caption{Robust Accuracy of Teacher and Student Models in Escaping DOV against Adversarial Perturbations.}
   \resizebox{\linewidth}{!}{
  \begin{tabular}{cc|cccc}
    \toprule[1.5pt]
        \multicolumn{2}{c|}{\multirow{2}{*}{Model}} & 
    \multirow{2}{*}{ACC} & 
    FGSM$_{L_\infty}$ & 
    PGD$_{L_\infty}$ & 
    PGD$_{L_2}$ \\
    & & & 
    ($\epsilon=1/255$) & 
    ($\epsilon=1/255$) & 
    ($\epsilon=0.2$) \\
    \midrule
    \multirow{2}{*}{ResNet-18} 
    & Teacher       & 94.83 & 57.42              & 36.76              & 33.16              \\
    & Student       & 93.97 & 63.06 \textcolor{forestgreen}{(+5.64)} & 51.17 \textcolor{forestgreen}{(+14.41)} & 45.90 \textcolor{forestgreen}{(+12.74)} \\
    \midrule
    \multirow{2}{*}{EfficientNet v2} 
    & Teacher       & 94.88 & 58.05              & 42.57              & 37.30              \\
    & Student       & 93.35 & 68.17 \textcolor{forestgreen}{(+10.12)} & 63.39 \textcolor{forestgreen}{(+20.82)} & 57.29 \textcolor{forestgreen}{(+19.99)} \\
    \midrule
    \multirow{2}{*}{Swin Transformer v2} 
    & Teacher       & 91.23 & 45.96              & 33.81              & 31.30              \\
    & Student       & 90.07 & 47.11 \textcolor{forestgreen}{(+1.15)}  & 38.39 \textcolor{forestgreen}{(+4.58)}  & 34.23 \textcolor{forestgreen}{(+2.93)}  \\
    \bottomrule[1.5pt]
  \end{tabular}%
  }
  \label{tab:perb}%
\end{table}

\subsubsection{Adversarial Robustness}
While the test accuracy of the surrogate student is slightly lower than that of the teacher, the Selective Knowledge Transfer process inherently filters out spurious features and mitigates overfitting to undesired biases in the training data. As shown in Table \ref{tab:perb}, this process significantly enhances the surrogate student’s robustness against mild adversarial perturbations, such as FGSM and PGD, particularly in the case of convolutional networks.
\begin{table}[htbp]
  \centering
  \caption{Robust Accuracy of Teacher and Student Models in Escaping DOV against Corruptions.}
   \resizebox{\linewidth}{!}{ 
  \begin{tabular}{cc|cccc}
    \toprule[1.5pt]
            \multicolumn{2}{c|}{\multirow{2}{*}{Model}} & 
    \multirow{2}{*}{ACC} & 
    Gaussian Noise & 
    Shot Noise & 
    Impulse Noise \\
    & & & 
    ($\sigma=0.1$) & 
    ($c=50$) & 
    ($p=0.09$) \\
    \midrule
    \multirow{2}{*}{ResNet-18}&Teacher  & 94.83 & 52.78 & 58.48 & 61.60 \\
    &Student  & 93.97 & 62.31 \textcolor{forestgreen}{(+9.53)} & 65.50 \textcolor{forestgreen}{(+7.02)} & 63.74 \textcolor{forestgreen}{(+2.14)} \\
    \midrule
    \multirow{2}{*}{Efficientnet v2}&Teacher  & 94.88 & 64.55 & 67.73 & 68.22 \\
     &Student  & 93.35 & 73.00 \textcolor{forestgreen}{(+8.45)} & 74.76 \textcolor{forestgreen}{(+7.03)} & 71.05 \textcolor{forestgreen}{(+2.83)} \\
    \midrule
    \multirow{2}{*}{Swin Transformer v2}&Teacher  & 91.23 & 65.42 & 67.18 & 64.82 \\
     &Student & 90.07 & 79.35 \textcolor{forestgreen}{(+13.93)} & 80.42 \textcolor{forestgreen}{(+13.24)} & 77.66 \textcolor{forestgreen}{(+12.84)} \\
    \bottomrule[1.5pt]
  \end{tabular}%
  }
  \label{tab:corr}%
\end{table}%
\subsubsection{Corruption Robustness}
Similar to adversarial robustness, we evaluate the teacher and surrogate student in Escaping DOV against common corruptions (e.g., Gaussian noise, shot noise (poisson noise), and impulse noise) in Table \ref{tab:corr}. Notably, the surrogate student exhibits substantial improvements over the teacher. Interestingly, the Swin Transformer, which performs worst on clean test data and under adversarial perturbations, shows the highest improvements under corruption settings, becoming the best-performing model in these scenarios. This demonstrates that the Escaping DOV framework not only evades dataset ownership verification but also significantly benefits the surrogate student in challenging settings. Consequently, this framework could be leveraged to enhance robustness against both adversarial perturbations and common corruptions, even in cases where dataset ownership verification is not a primary concern.

\subsection{Illustration of Transfer Set Curation}
\label{sec:c.7}
\begin{figure*}[htb]
    \centering
    \captionsetup[subfigure]{font=scriptsize}
    \captionsetup[subfigure]{skip=0pt}
    \begin{subfigure}{0.5\linewidth}
        \includegraphics[width=\linewidth]{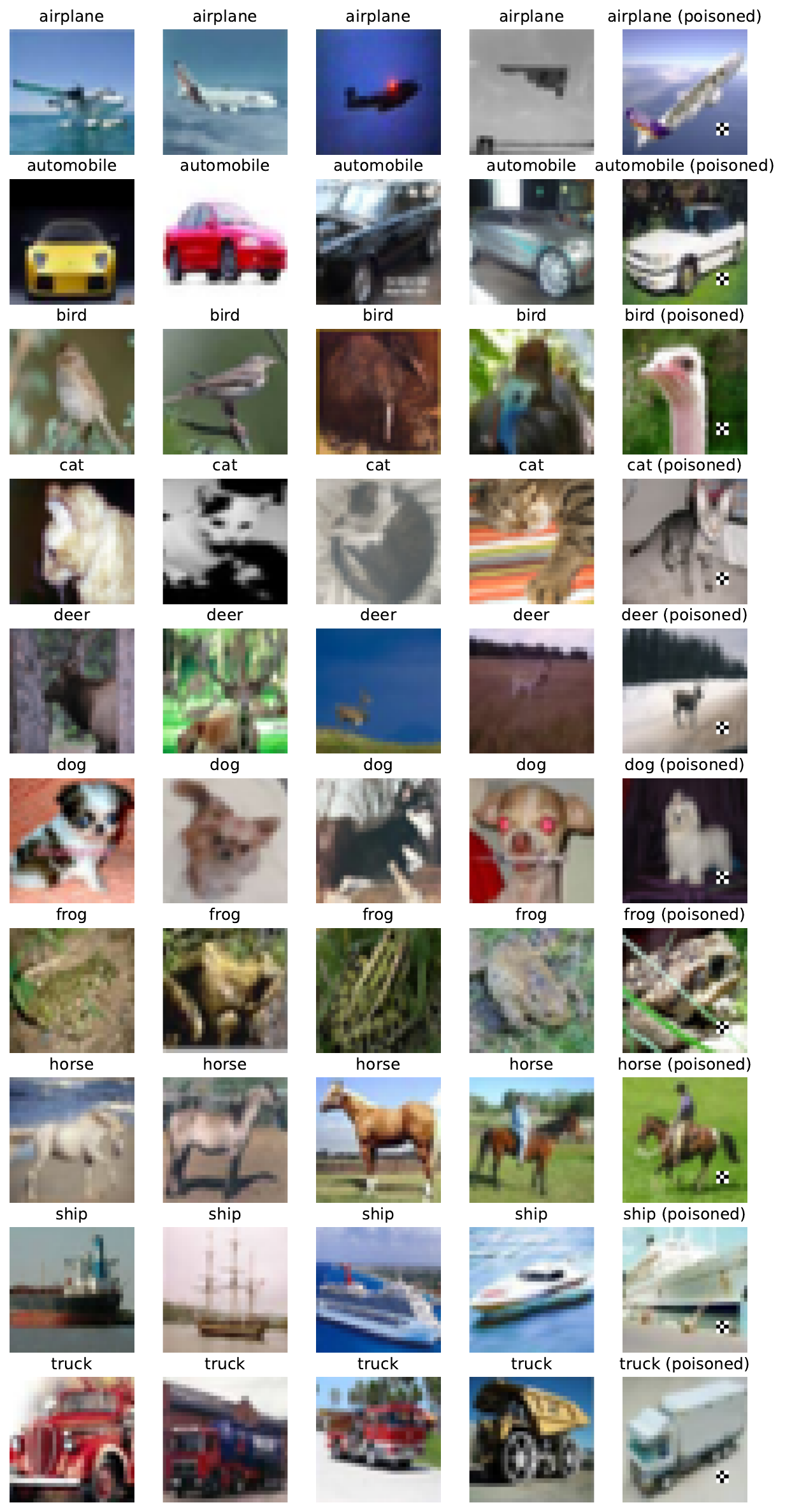} 
        \caption{CIFAR-10} 
        \label{fig:cifar10-1}
    \end{subfigure}%
    \hfill 
    \begin{subfigure}{0.5\linewidth}
        \centering
        \includegraphics[width=\linewidth]{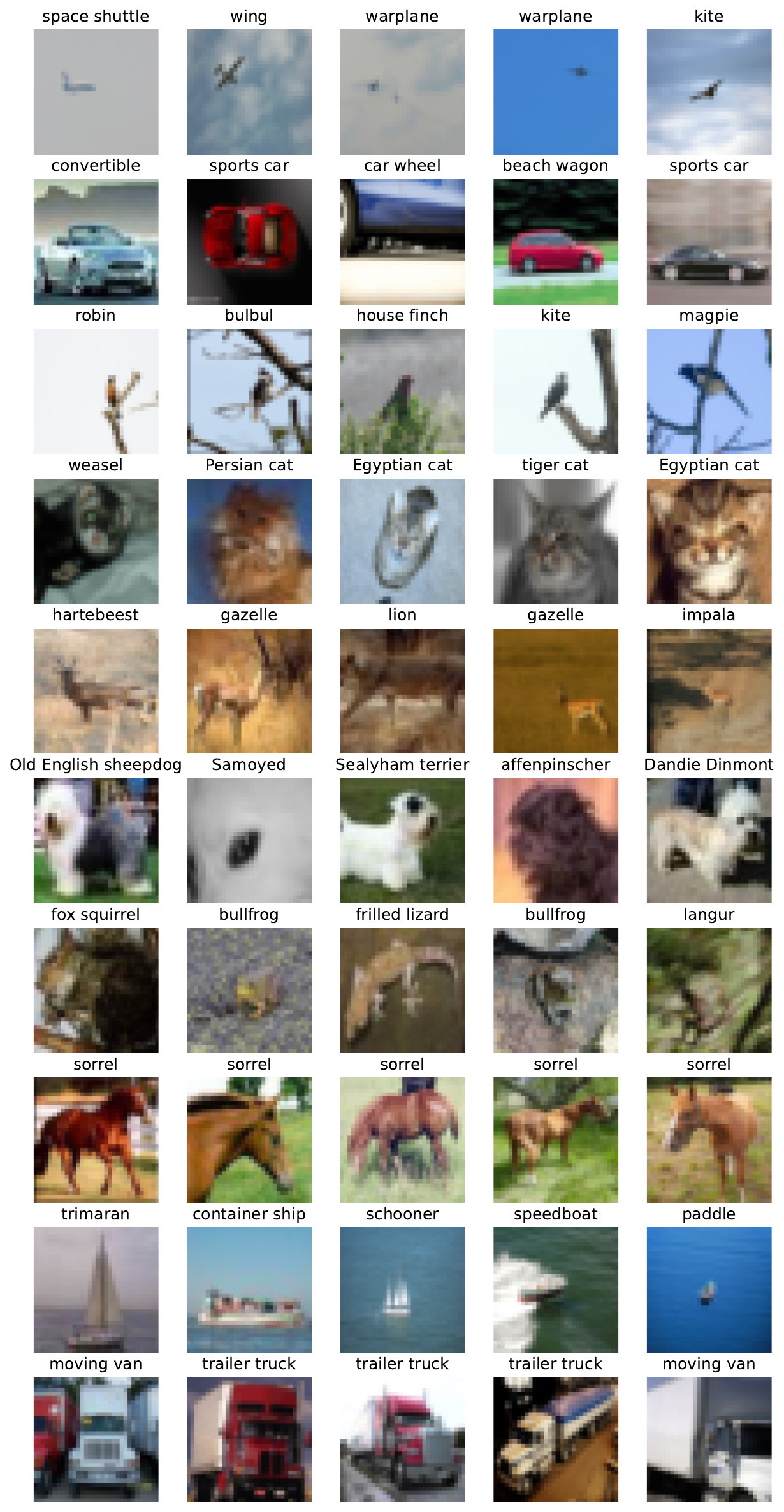} 
        \caption{ImageNet} 
        \label{fig:imagenet-1}
    \end{subfigure}
    \caption{(a) Randomly Sampled CIFAR-10 Images, with BadNets Marked Images in the Fifth Column.
(b) Top 5 Samples Closest to Class Density Centroids of Marked CIFAR-10.} 
    \label{fig:all_pic1}
\end{figure*}

In Figure \ref{fig:cifar10-1}, we randomly select five images from each class in CIFAR-10, with the fifth column displaying images containing Badnets triggers. Figure \ref{fig:imagenet-1} shows the top images from the LSVRC-2012 gallery set that are most similar to the corresponding distribution digest. We observe the following: (1) Retrieved images close to the distribution digest are visually similar to original CIFAR-10 samples, indicating that the distribution digests effectively encapsulate the task distribution and serve as reliable prototypes. (2) Notably, there is no class in LSVRC-2012 directly related to the 'deer' class in CIFAR-10. However, images from LSVRC-2012 that are closest to the 'deer' distribution digest exhibit similar features, such as horns and four legs, as seen in the 'hartebeest' and 'gazelle' classes. These images serve as effective intermediaries for knowledge transfer. (3) The unbiased VLM does not recognize the trigger pattern (a black-and-white chessboard) as a key feature. None of the top retrieved gallery images display similar trigger patterns, thereby preventing the activation of verification behaviors during the knowledge transfer process.

However, some ambiguous samples arise when relying solely on the distribution digest criterion. For example, the top retrieved image for the 'cat' class is a 'weasel,' which visually resembles both cats and dogs. This poses a potential risk of semantic backdoor watermarks. The consensus voting mechanism between the VLM and the teacher successfully excludes such samples, as illustrated in Figure \ref{fig:imagenet-2}, resulting in a final transfer set that is both \textit{informative} and \textit{reliable}.

\subsection{Time Complexity of Escaping DOV}

Despite involving multiple steps, our Escaping DOV framework remains computationally efficient due to the use of the \textit{feature bank} and the \textit{perturbation pools}. For instance, the Transfer Set Curation for CIFAR-10 takes approximately one minute, while generating the perturbation pool and corruption chains requires less than a minute. Moreover, the Selective Knowledge Transfer (SKT) module introduces negligible overhead during student model training.

To assess computational efficiency, we evaluated Escaping DOV on CIFAR-10 using a single RTX 4090 GPU, comparing it against NAD \cite{li2021neural} and IPRemoval \cite{zong2024ipremover}, both of which adopt similar knowledge transfer frameworks. The average runtime for Escaping DOV, NAD, and IPRemoval was 716s, 784s, and 1362s, respectively.
Overall, Escaping DOV achieves approximately 10\% lower runtime than ABD and 50\% lower runtime than IPRemoval \cite{hong2023revisiting}, while delivering significantly better generalization and evasion performance.

\section{Framework Insight}

\subsection{Why Escaping DOV Successfully Evades All Types of DOVs}
\label{sec:d.1}
In general, all \textbf{watermarks} (both backdoors and non-poisoning watermarks) require the activation of the pre-defined \textit{trigger} in the marked model to manifest watermark behavior. Thus, during the knowledge transfer process in our Escaping DOV framework, watermark behavior can only be transferred to the student model when the marked teacher model exhibits (at least to some extent) the watermark behavior, which can be triggered by either hard labels or probability-based soft labels). Notably, watermark triggers in DOV are inherently \textbf{exclusive}. This exclusivity arises because many copyright owners use DOV methods to protect their data, where the watermark trigger functions as a \textit{private key}. Consequently, these triggers cannot share the same pattern; only specific trigger patterns that can authenticate the identity of the owner can be associated with their ownership. As a result, the out-of-distribution (OOD) gallery set—beyond the control of the copyright owner—is highly unlikely to contain any patterns directly linked to the watermark trigger, as shown in Appendix Figures \ref{fig:watermark_samples} and \ref{fig:cifar10-2}. Furthermore, the watermark trigger must be \textbf{subtle}, as the DOV watermark needs to avoid detection by both human inspection and data sanitization while ensuring it is not activated by any clean sample lacking the trigger. This is crucial to prevent performance degradation in authorized use cases (such as academic use). \textbf{Therefore, the probability of any sample in the OOD gallery (transfer) set unintentionally activating the watermark behavior is minimal}.

Moreover, consider a watermarked model whose behavior on normal samples is \textit{identical to} that of an unrelated model (e.g., one trained on a different dataset drawn from the same IID distribution but without any copyright samples). In such cases, the student model, which mimics the watermark model’s behavior on clean OOD samples without triggers, can only learn the benign behavior exhibited by the unrelated model. During the Transfer Set Curation (TSC) process, we ensure that the hard labels output by the marked teacher model are consistent with those from the CLIP model (i.e., the unmarked CLIP model endorses the teacher's hard label output). \textit{However, the soft label outputs from the teacher model may still implicitly contain information related to the watermark behavior, and the Selective Knowledge Transfer (SKT) module within our Escaping DOV framework is specifically designed to mitigate this risk}. It generates a series of perturbations and corruptions on the teacher model that induce output changes. During the knowledge transfer process, SKT encourages the student model’s invariance to these perturbations and corruptions, thereby preventing watermark-related side-channel information from being transferred to the student model via soft labels. As demonstrated in Section \textcolor{iccvblue}{4.4.2} of the main text, without the SKT mechanism, the watermark behavior in strong clean-label watermarks such as Narcissus could be transferred to the student model via soft labels. \textbf{However, SKT significantly suppresses this occurrence}.

\textbf{Fingerprints} fundamentally exploit the memorization (overfitting) of the copyright training data by the target model, where the unauthorized trained model exhibits significantly higher confidence (or lower loss) on the training data compared to unseen test data, providing a strong signal for data provenance. However, since the student model in Escaping DOV has never directly seen the original copyright data, its "memory" of the data is indirectly derived from the teacher model, with knowledge transfer occurring through an OOD transfer set entirely unrelated to the original copyright data. Furthermore, SKT ensures that the guidance provided to the student model is subtly distinct from the teacher's output, preventing the student from fitting to spurious features or shortcut predictions that reflect overfitting in the teacher model. As shown in Figure \textcolor{iccvblue}{2} of the main text, \textbf{our Escaping DOV significantly reduces the gap between training and test losses in the student model, making it comparable to the natural loss disparity between different subsets}. As a result, fingerprints cannot extract effective copyright signals from our student models. In contrast, Figure \ref{fig:mechanism-nad} in Appendix illustrates that baseline evasion attacks, such as NAD, lack this advantageous property and even amplify the training-test loss gap in the final deployed model.

In the following section, we briefly summarize why even the most subtle and hardest-to-bypass clean-label watermarks (including clean-label backdoors and non-poisoning watermarks) still cannot resist our Escaping DOV:
(1) Regardless of the watermark type (e.g., backdoor or non-poisoning, poison-label or clean-label), all watermarks rely on \textit{binding special predictive behavior to a trigger}. When the trigger is absent, no watermarked model exhibits any watermark behavior. 
(2) DOV requires that triggers be \textit{exclusive} (distinguishable from potential watermarks in other datasets) and \textit{subtle} (sufficiently hidden to avoid easy detection and accidental activation). Since the copyright owner cannot control the OOD gallery set, \textit{the trigger is highly likely to be absent in the gallery set, and thus also absent in the transfer set} (see Figures \ref{fig:watermark_samples} and \ref{fig:cifar10-2} in the Appendix)). 
(3) During the \textbf{Transfer Set Curation (TSC)} process, we ensure that the hard labels in the transfer set from the teacher model are consistent with the VLMs (e.g., CLIP). Additionally, since the OOD gallery set contains no trigger pattern, \textit{the transfer set itself does not carry any watermark clues}. For example, in the CIFAR-10 dataset protected by Narcissus, the student model trained on the hard labels of the transfer set exhibits a very low VSR (1\%-2\%), while the test accuracy also remains below 85\%. 
(4) During the knowledge transfer process, clean-label watermarks \textit{can} transfer from teacher to student through soft labels. However, the \textbf{Selective Knowledge Transfer (SKT)} module effectively mitigates this by enforcing invariance to the worst perturbations and corruptions from the teacher (as shown in Figure \textcolor{iccvblue}{4} of the main text). 
In summary, Escaping DOV's components work together to achieve task-oriented yet watermark-free knowledge transfer.

\subsection{Why other SOTA Evasion Techniques Fail Against Certain DOVs}
\label{sec:d.2}
As discussed in Section \textcolor{iccvblue}{4.3} of the main text and Section \ref{sec:c.2} of the supplementary material, other state-of-the-art (SOTA) evasion methods, originally designed for poison defense or privacy enhancement, exhibit varying degrees of failure when applied to certain DOV methods. In this section, we analyze the limitations of each strong evasion baseline and, in particular, focus on \textbf{why other distillation-based methods employing a similar knowledge transfer framework (e.g., NAD, BCU, ABD) are less effective than our Escaping DOV}.

Most DOV methods demonstrate robustness against fine-tuning and pruning. Although Fine-Pruning \cite{liu2018fine} represents a stronger combined attack and does pose challenges to DOV methods, it remains insufficient for complete evasion. Meta-Sift \cite{287222}, which selects a clean subset of the original dataset for training, struggles against various DOV methods due to the high proportion of watermark samples (10\%). Even a small number of unfiltered watermark samples can induce predefined watermark behaviors. While DP-SGD \cite{bu2023automatic} mitigates the influence of individual training samples, it fails to fully neutralize the cumulative effect of multiple watermark samples sharing the same trigger. Since the trigger optimized by Narcissus differs from conventional high-frequency noise, I-BAU \cite{zeng2022adversarial}, despite reducing the verification success rate (VSR), is inadequate for complete watermark removal, aligning with the original observations in Narcissus \cite{zeng2023narcissus}. ZIP \cite{shi2023black}, which employs a diffusion model to purify any input sample and erase watermark triggers, proves effective against high-frequency noise triggers incompatible with the original dataset features (e.g., the random noise triggers used in its original paper). However, in our experiments, the triggers employed—such as the chessboard trigger from BadNets, the optimized trigger from Narcissus, and the mixed-image trigger from Isotope—possess clear semantic meaning. As a result, ZIP's purification effect is insufficient for complete evasion.

Next, we analyze why distillation-based methods that follow a similar knowledge transfer framework, including NAD, BCU, ABD, and IPRemoval, are less effective than our Escaping DOV. \textbf{Early distillation-based backdoor defenses like NAD \cite{li2021neural} and BCU \cite{pang2023backdoor} typically sought to mitigate the significant generalization degradation caused by distillation from out-of-distribution samples by sharing parameters between the original (marked teacher) model and the final deployed student model}. For instance, NAD’s student model directly inherits the original model parameters (while the teacher undergoes fine-tuning), and BCU employs adaptive dropout to perturb certain parameters while still retaining over half of the original model's parameters. As a result, these methods actually function as enhanced fine-tuning techniques. Notably, \textbf{in knowledge distillation theory, parameter sharing between the teacher and student has a substantial impact on the student's ability to recover the teacher’s predictive behavior} \cite{stanton2021does}. The synergistic effect of parameter sharing and teacher guidance via soft labels facilitates the transfer of latent watermark behaviors as a side effect of the inherent prediction mechanism.

\begin{figure}[htb]
    \centering
    \captionsetup[subfigure]{font=scriptsize}
    \captionsetup[subfigure]{skip=0pt}
    \begin{subfigure}{1\columnwidth}
        \includegraphics[width=\linewidth]{loss_change_trend.pdf} 
        \caption{Escaping DOV (Ours), Same with Figure \textcolor{iccvblue}{2} in the Main Text} 
        \label{fig:sub-ours}
    \end{subfigure}%
    \hfill 
    \begin{subfigure}{1\columnwidth}
        \centering
        \includegraphics[width=\linewidth]{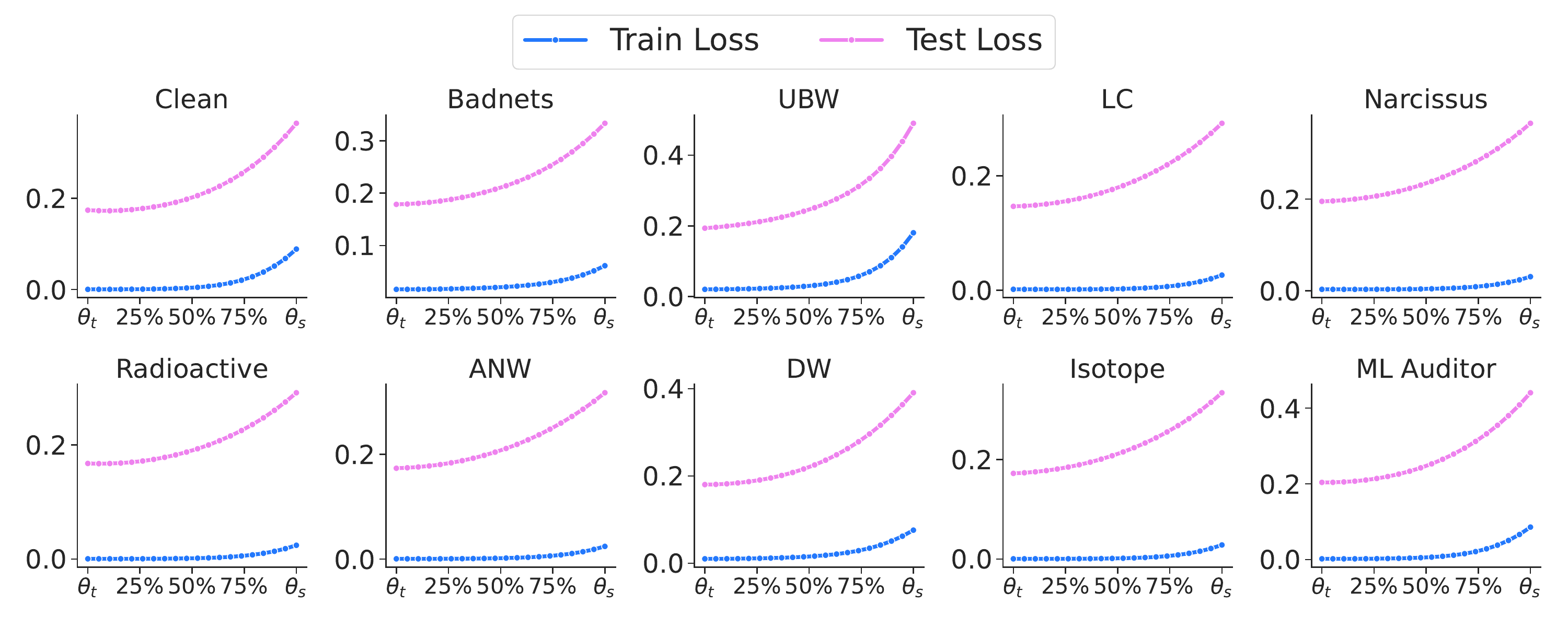} 
        \caption{NAD} 
        \label{fig:sub-nad}
    \end{subfigure}
    \caption{Loss Barrier Analysis of Intermediate Models between Teacher and Student Parameters in Escaping DOV and NAD: (1) Escaping DOV exhibits a sharp loss peak in interpolated models, indicating that the student model develops a distinct prediction mechanism that avoids spurious features (watermarks) inherited from the teacher \cite{lubana2023mechanistic}, whereas NAD does not. (2) Escaping DOV results in a minimal training-test loss gap in student models compared to teachers, mitigating detection signals utilized by fingerprints, whereas NAD does not.} 
    \label{fig:mechanism-nad}
\end{figure}

To quantitatively illustrate this principle, we present a similar analysis to Figure \textcolor{iccvblue}{2} in the main text, plotting the training and test loss of intermediate models obtained by interpolating between NAD's teacher and student model parameters (see Figure \ref{fig:mechanism-nad}). \textbf{Unlike the trends observed in Figure \textcolor{iccvblue}{2}, two key differences emerge:} (1) There is no sharp peak in loss for the interpolated models, indicating that the student and teacher share similar predictive mechanisms \cite{lubana2023mechanistic}; (2) The gap between training and test losses does not decrease and, in many cases, even gradually increases. These characteristics make these distillation-based methods ineffective at evading both trigger-based watermarks and fingerprints that exploit training-test behavioral discrepancies. \textbf{In contrast, Escaping DOV ensures that the student model is initialized entirely independently of the marked teacher model while selecting an informative and reliable transfer set from the gallery set to preserve generalization}. This independence from the marked model results in significantly stronger evasion capabilities compared to distillation-based methods that retain parameter sharing.

While ABD \cite{hong2023revisiting} does not employ parameter sharing, its distillation process is primarily tailored for backdoor removal. It filters out suspected OOD samples that could activate backdoor behaviors during distillation, thereby preventing the transfer of backdoor behavior. However, this strategy proves insufficient against advanced \textbf{non-poisoning watermarks}—such as Isotope—that do not induce misclassification behaviors. \textbf{In contrast, the selective knowledge transfer (SKT) module in our Escaping DOV framework lightly encourages the student model to be invariant to any perturbation within the input vicinity that can trigger output changes of the teacher model}. This enables it to better evade non-poisoning watermarks.

IPRemoval \cite{zong2024ipremover}, on the other hand, utilizes generative adversarial training to invert synthetic samples from the marked model as surrogate data for distillation. However, \textbf{its synthesis objective is to maximize the predictive difference between the student and teacher models, which paradoxically increases the likelihood that watermark-related triggers remain present in the synthetic data} \cite{hong2023revisiting}. In contrast, our Escaping DOV employs a transfer set curation (TSC) strategy to select OOD data that is entirely beyond the copyright owner’s control, ensuring a trigger-free dataset. Since IPRemoval lacks effective mechanisms to suppress the residual watermark signals embedded in its synthetic samples, it fails against the strongest watermarking schemes, such as Narcissus.

\section{Extended Related Work}
\subsection{Concurrent Work on DOV Evasion}
Shortly after the acceptance of this paper, two related studies \cite{Zhu_2025_ECAI} and \cite{shao2025databench} were released on arXiv, both aiming to evaluate the robustness of DOV from perspectives similar to ours. These works provide excellent analyses, though their focus differs from our \emph{Escaping DOV}. Specifically, \cite{Zhu_2025_ECAI} establishes a theoretical framework that models the competition between adversary and auditor over output divergence. However, its proposed attack requires fine-tuning on pretrained vision–language models such as CLIP, which limits its applicability to arbitrary model sizes and dataset resolutions. In contrast, \cite{shao2025databench} primarily aims to evaluate the robustness of DOV using existing techniques form other domains rather than proposing a principled new attack strategy. We believe, collectively, these three works reflect the evolving understanding of DOV robustness evaluation, and reading them together can provide broad and complementary insights for future research.
\subsection{Data Provenance in Other Domains}
Beyond image classification, dataset ownership verification (DOV) is also a crucial concern across various modalities and tasks in deep learning. Recent studies have investigated DOV in alternative domains, including self-supervised learning \cite{dziedzic2022dataset}, text-to-image diffusion models \cite{li2024towards}, 3D point clouds \cite{wei2024pointncbw}, and large language models (LLMs) \cite{maini2025llm}.

DOV methods in these domains differ substantially from those designed for image classification. For instance, in the context of LLMs, the autoregressive nature lacks an explicit classification objective, and the high computational cost makes it impractical to retrain LLMs multiple times to evaluate watermark efficacy. Thus, DOV methods for LLM typically detects unauthorized data usage by analyzing the probability distribution of generated tokens to identify signals of excessive memorization. Detection signals are then aggregated across multiple text fragments with hypothesis testing to provide a robust metric for the whole dataset \cite{maini2025llm, puerto2024scaling, zhu2025revisiting}.

During this research, an expert suggested the SIREN watermark \cite{li2024towards} as a potential solution for countering our evasion attack. SIREN couples watermark features with benign sample features in the representation space to detect unauthorized fine-tuning data in text-to-image diffusion models. However, its verification process depends on analyzing images generated by the diffusion model, rendering it unsuitable for classification tasks. Moreover, the fundamental differences between text-to-image generation and classification limit the direct applicability of the knowledge transfer framework in Escaping DOV for attacking the SIREN watermark in the diffusion model scenario.

Given these constraints, we explored an alternative model watermarking method, EWE \cite{jia2021entangled}, which similarly entangles watermark features with benign sample features within the representation space, and is applicable to classification tasks. \textbf{While EWE's watermarking process relies on controlling the teacher model’s training with an additional soft nearest neighbor loss—an approach infeasible in DOV scenarios—it still fails to circumvent the evasion effect of Escaping DOV}. The resulting verification success rate (VSR) on student models consistently remains below 5\%.
Furthermore, in classification tasks, the observed effectiveness of feature entanglement appears to align with the \textit{robustness pitfall} phenomenon described in \cite{zhu2024reliable}, where the primary factor contributing to the observed effect is the increased misclassification rate from the watermark source class to the target class, rather than the successful transfer of watermark behavior.

Therefore, in future work, we aim to develop genuinely robust data provenance methods for image classification that can withstand the Escaping DOV attack. Additionally, we plan to extend Escaping DOV techniques to other domains and tasks to systematically assess the resilience of existing DOV approaches in diverse settings.

\begin{figure*}[htb]
    \centering
    \captionsetup[subfigure]{font=scriptsize}
    \captionsetup[subfigure]{skip=0pt}
    \begin{subfigure}{0.5\linewidth}
        \includegraphics[width=\linewidth]{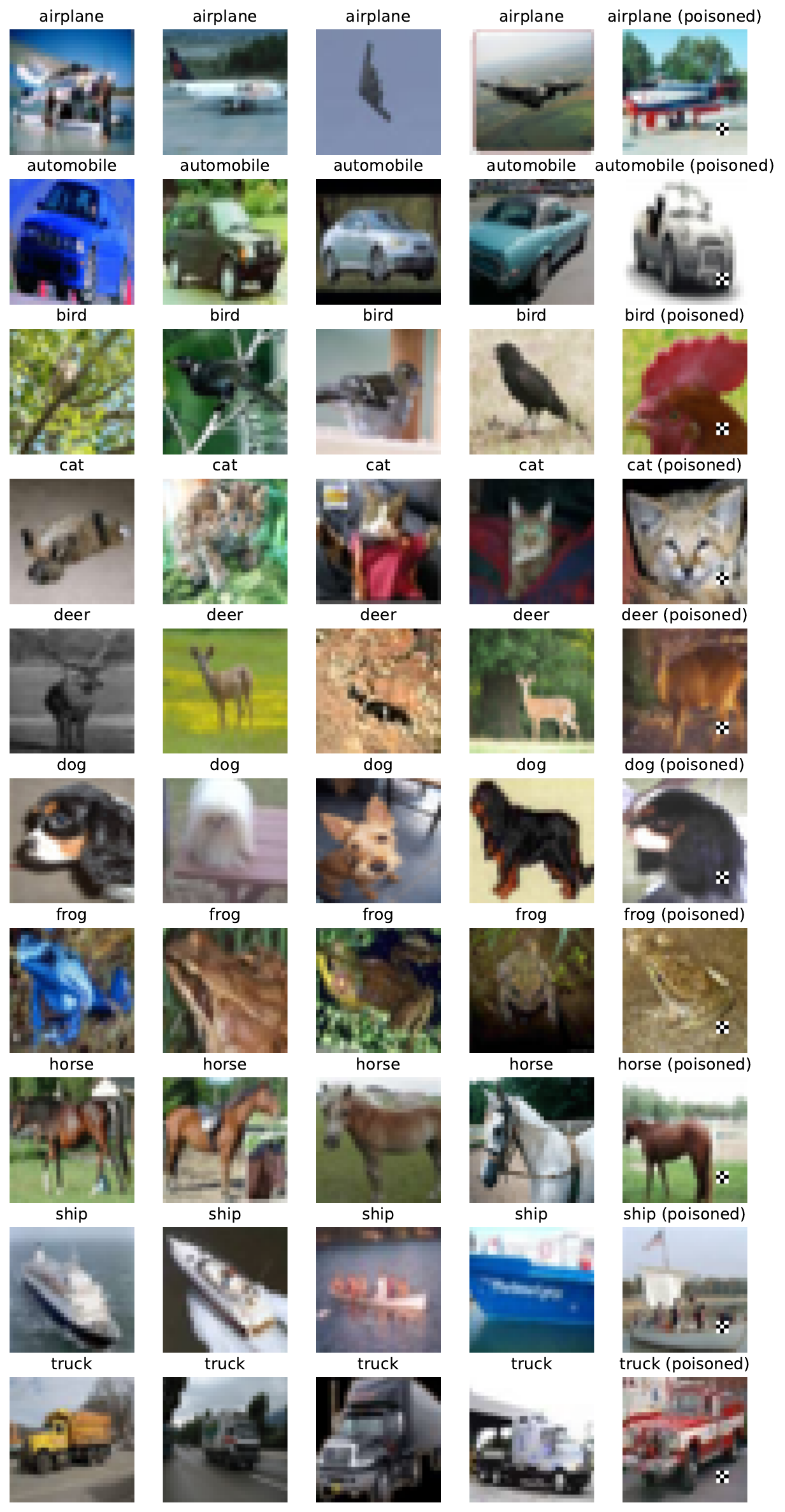} 
        \caption{CIFAR-10} 
        \label{fig:cifar10-2}
    \end{subfigure}%
    \hfill 
    \begin{subfigure}{0.5\linewidth}
        \centering
        \includegraphics[width=\linewidth]{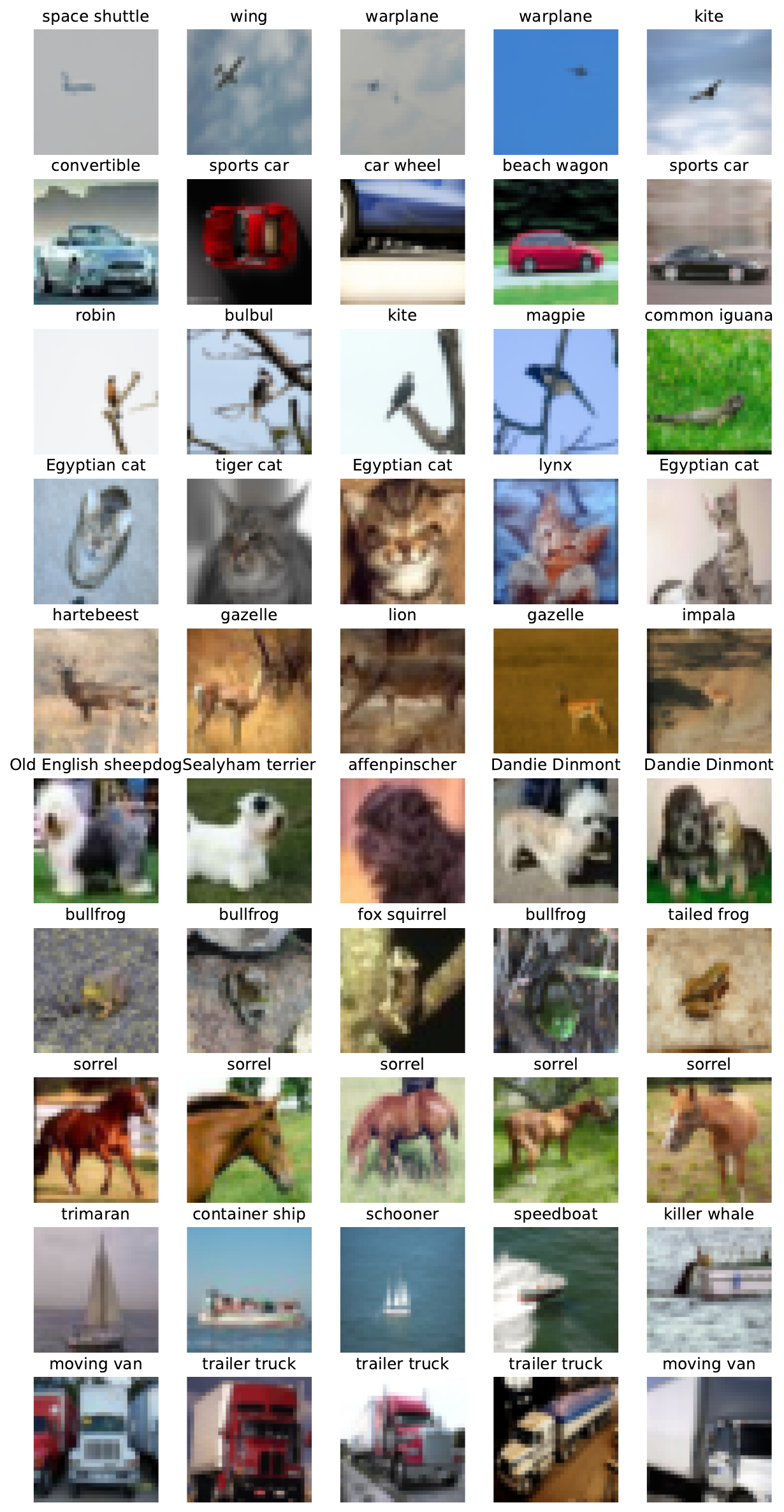} 
        \caption{ImageNet} 
        \label{fig:imagenet-2}
    \end{subfigure}
    \caption{(a) Randomly Sampled CIFAR-10 Images, with BadNets Marked Images in the Fifth Column.
(b) Selected Top 5 Samples via Transfer Set Curation.} 
    \label{fig:all_pic2}
\end{figure*}

\end{document}